%% file: main.tex
\documentclass[sigconf]{acmart}

\AtBeginDocument{%
  }

\setcopyright{acmcopyright}
\copyrightyear{2025}
\acmYear{2025}
\setcopyright{acmlicensed}\acmConference[KDD '26]{Proceedings of the 32th ACM SIGKDD Conference on Knowledge Discovery and Data Mining}{August 9--13, 2026}{Jeju, ON, Korea}
\acmBooktitle{Proceedings of the 32th ACM SIGKDD Conference on Knowledge Discovery and Data Mining (KDD '26), August 9--13, 2026, Jeju, ON, Korea}
\acmISBN{978-1-4503-XXXX-X/18/06}

\usepackage{amsmath}
\usepackage{bm}
\usepackage{enumitem}
\usepackage{tikz}
\usepackage{subcaption}
\usepackage{makecell}
\usepackage[ruled]{algorithm2e}
\usepackage{siunitx}
\usepackage{amsthm}
\newtheorem{observation}{Observation}

\SetKwInput{KwInput}{Input}
\SetKwInput{KwOutput}{Output}
\begin{document}

\title{SolarBoost: Distributed Photovoltaic Power Forecasting Amid Time-varying Grid Capacity}


\author{Linyuan Geng}
\email{genglinyuan.gly@alibaba-inc.com}
\affiliation{\institution{DAMO Academy, Alibaba Group}\city{Hangzhou}\country{China}}
\authornote{Equal Contribution}

\author{Linxiao Yang}
\email{linxiao.ylx@alibaba-inc.com}
\affiliation{\institution{DAMO Academy, Alibaba Group}\city{Hangzhou}\country{China}}\authornotemark[1]

\author{Xinyue Gu}
\email{guxinyue.gxy@alibaba-inc.com}
\affiliation{\institution{DAMO Academy, Alibaba Group}\city{Hangzhou}\country{China}}
\authornotemark[1]

\author{Liang Sun}
\email{liang.sun@alibaba-inc.com}
\affiliation{\institution{DAMO Academy, Alibaba Group}\city{Hangzhou}\country{China}}
 

\renewcommand{\shortauthors}{Linyuan Geng, Linxiao Yang, Xinyue Gu \& Liang Sun}

\begin{abstract}
This paper presents SolarBoost, a novel approach for forecasting power output in distributed photovoltaic (DPV) systems. While existing centralized photovoltaic (CPV) methods are able to precisely model output dependencies due to uniformity,  it is difficult to apply such techniques to DPV systems, as DPVs face challenges such as  missing grid-level data, temporal shifts in installed capacity, geographic variability, and panel diversity. SolarBoost overcomes these challenges by modeling aggregated power output as a composite of output from small grids, where each grid output is modeled using a unit output function multiplied by its capacity. This approach decouples the homogeneous unit output function from dynamic capacity for accurate prediction. Efficient algorithms over an upper-bound approximation are proposed to overcome computational bottlenecks in loss functions. We demonstrate the superiority of grid-level modeling via theoretical analysis and experiments. SolarBoost has been validated through deployment across various cities in China, significantly reducing potential losses and provides valuable insights for the operation of power grids. The code for this work is available at \href{https://github.com/DAMO-DI-ML/SolarBoost}{https://github.com/DAMO-DI-ML/SolarBoost}.
\end{abstract}

\maketitle
\input{1.introduction}

\input{2.background}
\input{3.method}

\input{4.theoretical_analysis}
\input{5.experiment}

\input{6.deployment}
\input{7.conclusion}

\newpage
\bibliographystyle{ACM-Reference-Format}
\bibliography{11.references}



\clearpage
\appendix
\input{10.appendix}
\input{2plus.related_work}

\end{document}

%% file: 1.introduction.tex
\section{Introduction}
\label{sec:intro}
Photovoltaic (PV) power converts sunlight directly into electricity using PV panel arrays, which is considered cost-effective, secure, and environmentally friendly. Consequently, PV generation is rapidly expanding, now accounting for 5.49\% of global electricity in 2023, marking a 35.9\% year-on-year increase in its share~\cite{ei2024energy}. There are two types of PV systems. Centralized PV (CPV): solar farms connected to the main power grids. Distributed PV (DPV): small installations on rooftops or in local communities. DPV systems are easy to install, require little maintenance and can utilize unused places such as rooftops or walls, making up 41\% of global PV capacity~\cite{pv2024} and helping make electricity available, clean and affordable for everyone.

Accurate forecasting of PV power output is essential for maintaining grid stability and maximizing the use of renewable energy. If solar power output is overestimated or underestimated, it can disrupt the balance between electricity supply and demand, sometimes leading to equipment overloads, forced curtailment, or even blackouts. For example, in August 2020, California experienced its largest rolling blackout since the 2001 energy crisis, due to miscalculations in balancing energy generation and demand~\cite{california}. Another example is a publicly listed Chinese wind power company~\cite{windcorp}, which reported that in the first half of 2023, 6,440 million kilowatt-hours, or 9.1\% of its generated power, were curtailed due to inadequate management of energy absorption. This is equivalent to 212,000 tonnes of standard coal waste, causing a direct economic loss over 42 million USD to this single company.

A substantial body of research exists on solar power forecasting~\cite{hong2020energy}, typically relying on weather predictions (such as irradiance and cloud cover) as input. Unfortunately, the majority of solar forecasting methods have been developed for CPV systems~\cite{chu2024review,rahdan2024distributed}, which have relatively uniform characteristics. In contrast, forecasting for DPV systems faces additional challenges, including:
\begin{enumerate}[leftmargin=*, nosep]
    \item \label{enu:missingdata}\textbf{Limited and Noisy Data:} Data from individual households or small-scale installations is often inconsistent, incomplete, or outdated, leading to difficulty in developing reliable models.
    \item \label{enu:shift}\textbf{Temporal Scale Shifts:} The capacity of DPV installations changes over time, introducing concept drift issues~\cite{AGRAHARI20229523}. We will dive into this issue in Section~\ref{subsec:grid_capacity}.
    \item  \label{enu:geo}\textbf{Geographical Diversity:} 
     DPV systems are widely scattered, experiencing different weather conditions. The approach that aggregates meteorological data to predict overall output as a single entity loses accuracy. Conversely, the method that predicts power data at consumer resolution is also less feasible due to limited and noisy data issues, as well as computational cost~\cite{rahdan2024distributed}.
    \item \label{enu:panel}\textbf{Equipment Diversity:} Unlike CPV systems with standardized equipment, DPV systems feature a mix of panel types, orientations and maintenance statuses, making it hard for consistent modeling.  
\end{enumerate}

To address these challenges of DPV power forecasting, we propose  \textbf{SolarBoost}, a novel boosting-based approach for modeling DPV power output. SolarBoost predicts total power by decomposing the area into small spatial grids and inferring two quantities for each grid: (1) a generalized (dynamic) grid-level capacity and (2) a per-unit-capacity output function that links weather to output. By decoupling capacity from the unit output function, our model can adjust to capacity changes with a consistent output function, thus improving forecast accuracy.

Here, ``grid'' refers to a spatial unit, distinct  the electricity power grid. Unless otherwise specified, the term ``grid'' in the following text refers to spatial grid points.

The main contributions of this study are:
\begin{enumerate}[leftmargin=*, nosep]
    \item \textbf{Capacity Modeling}: We develop a framework to explicitly model dynamic, heterogeneous grid-level generalized-capacities, which serves as a multiplier for the output capturing all heterogeneity, including capacity changes (challenge~\ref{enu:shift}), panel diversity (challenge~\ref{enu:panel}) and missing grid-level information (challenge~\ref{enu:missingdata}). 
    \item \textbf{Consistent Unit Output Function}: We introduce a consistent unit output function, ensuring stable weather-to-output modeling even as capacities evolve (challenge~\ref{enu:shift}).  Thus we can use broad data without sacrificing granularity (challenge~\ref{enu:geo}).
    \item \textbf{Efficient Optimization Framework on Boosting Trees}: We present an efficient, scalable algorithm based on boosting trees with upper-bound approximations to tackle the computational difficulties in the loss function involving nested summations. 
    \item \textbf{Theoretical Proof for Accuracy}: We provide  theoretical analysis that substantiates the superiority of grid-based models over aggregated approaches. This analysis also demonstrates the adequacy of  rough generalized-capacity estimations for accurate forecasts. 
    
    \item \textbf{Deployment}: SolarBoost has been  deployed in various cities of China, showing its efficacy and adaptability in real-world DPV forecasting  across different geographical  contexts.
\end{enumerate}

Extensive business experience and prior research have shown that ensemble trees, particularly boosting frameworks like XGBoost and LightGBM, perform well with medium-sized tabular datasets and have outperformed deep learning methods in PV forecasting tasks~\cite{hong2020energy,rahimi2023comprehensive}. This motivates our exploration of innovations based on the gradient-boosted trees paradigm.



%% file: 2.background.tex
\vspace{-.2cm}
\section{Background}
\label{sec:background}
\subsection{DPV Power Forecasting as a Regression Task} 

DPV power forecasting aims to predict future solar energy production over a region. In our deployment, the utility company provides bidirectional smart meters that record both the gross PV output that is produced by the PV system before any is used, and the net output that remains after household consumption. Throughout this paper, the forecasting target is defined as the aggregated gross PV power output (prior to any household consumption or behind-the-meter use), which is directly measured and available in our dataset. We do not model or predict household self-consumption. In real practice, all input data are obtained daily from a weather data supplier and all capacity data are obtained monthly from official power supply authorities, ensuring full data reliability.

At its core, the DPV power forecasting task answers: \textit{How much solar electricity will an area generate tomorrow, given predicted weather conditions and other features?}


\noindent\textbf{Key Concept: Installed Capacity. } 
The maximum power that a DPV system can generate is determined by its installed capacity--the total power generation capability of all solar panels in the system (``how big the system is''). Think of this as the system's ``size'':
\begin{itemize}[leftmargin = .2cm]
    \item Is physically determined by the number/type of photovoltaic panels,
    \item Is measured in kilowatts (kW) or megawatts (MW),
    \item Represents the output under ideal sunlight conditions (no clouds, optimal panel angle).
\end{itemize}

For example, in perfect sunlight, a 1MW system could theoretically produce 1MWh of electricity per hour. In practice, the real output is lower and changes based on the weather, but it is always capped by this installed capacity.


\noindent\textbf{Task Description. } 
Our models predict power output $y$ over $T$ time steps, using:
\begin{itemize}[leftmargin = .2cm]
    \item Weather forecasts: solar irradiance, cloud cover, temperature, etc.,
    \item Time-related factors: seasonal effect, time-of-day patterns, etc.
\end{itemize}
These inputs are structured as a $T\times D\times K$ tensor: $T$ time steps (such as days or 15-min intervals),  $D$-dimensional input variables (such as irradiance, cloud cover and temperature), and  $K$ spatial grid points (dividing the city into small geographic areas, each called a ``grid'').

\noindent\textbf{Distinct Properties. } While the task resembles standard temporal regression, it diverges from standard settings in two critical aspects: 1) the \textit{time-varying installed capacity} that scales the power over time, and 2) the \textit{many-to-one mapping} from sets of grid-level features to the overall output for a region: in other words, data from many local areas must be combined to produce a single prediction.  Since the grid-level output and capacities are not available, most prior methods cannot take full advantage of grid-level inputs. 

\vspace{-.2cm}
\subsection{Impact of Changing Capacity}
\begin{figure}[ht]
    \centering
    \includegraphics[width=\linewidth]{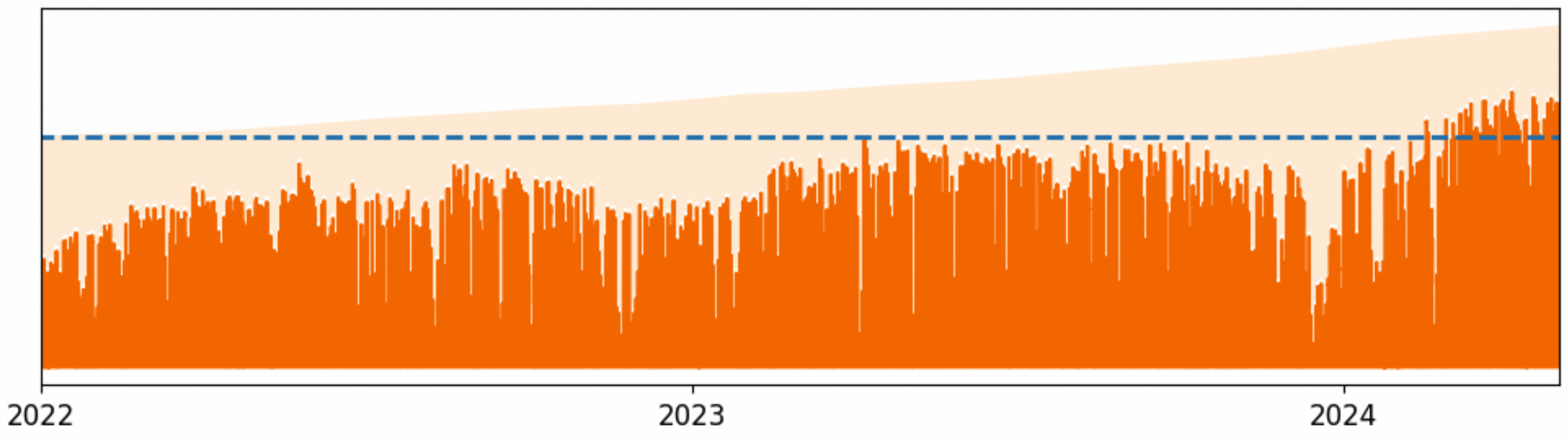}
    \vspace{-.6cm}
    \caption{Relationship between installed capacity and power output. Dark orange curve: actual DPV power output in an eastern Chinese city. Light orange shade: the installed capacity of this city. Dashed blue line: initial capacity for reference. Capacity expansion lifts the potential output ceiling, with daily output fluctuations governed by weather.}
    \label{fig:growing_capacity}
    \vspace{-.4cm}
\end{figure}

DPV capacity evolves over time. As displayed in Figure~\ref{fig:growing_capacity}, actual power (dark orange curve) fluctuates with weather but follows an upward trend dictated by the city-level capacity expansion (light orange band).  This creates an intrinsic \textit{scaling law}: the same weather conditions produce proportionally higher output as capacity grows. Theoretically, each additional panel increases the total capacity linearly, scaling potential output proportionally.

\begin{observation}\label{obs:multiplier}
\textbf{Capacity as a Multiplicative Factor}. 
For fixed weather conditions $\bm{x}$, the power output $y_t$ at time $t$ scales linearly with its installed capacity $c_t$:
\begin{equation}
    y_t = c_t \cdot f(\bm{x}),
\end{equation}
where $f(\cdot)$ converts weather features to per-unit-capacity output of photovoltaic power, assuming ideal conditions where photovoltaic conversion efficiency does not decay over time. Capacity changes \textit{rescale} the weather-to-power mapping. This means that if all other conditions are held constant, doubling the installed capacity will double the expected maximum output.
\end{observation}

Section \ref{sec:Problem} will consider more realistic scenarios, addressing potential decays in conversion efficiency in $f$.

\subsection{Limitations of Ignoring Grid Capacity}
\label{subsec:grid_capacity}
Current DPV forecasting methods primarily fall into two paradigms:

\noindent\textbf{Averaging geographical inputs:} Grid-level inputs are averaged to obtain $T\times D$ dimensions. In this way, the many-to-one mapping in location is converted into one-to-one, allowing standard regression modeling. However, this approach can be limiting in DPV systems due to variations in weather conditions across grids.

\noindent\textbf{Flattening geographical inputs:} This paradigm concatenates $D$-dimensional features from all $K$ grids  into flattened $D\cdot K$ vectors. While preserving raw information, this strategy treats all grids with the same static sensitivity.

\begin{figure}
\scriptsize
\noindent\hspace*{-0.5em} 

\begin{minipage}[t][][t]{0.1\linewidth}
    \vspace*{0pt}
    \rotatebox[origin=l]{90}{Grid 1\quad }
    
    \rotatebox[origin=l]{90}{Grid 2\quad }
    
    \rotatebox[origin=l]{90}{Grid 3\quad }
    
    \rotatebox[origin=l]{90}{$\ldots$}
    
    \rotatebox[origin=l]{90}{Grid K$\ \ $}
\end{minipage}%
\hspace{-.8cm}
\begin{minipage}[t][][t]{0.4\linewidth}
    \vspace*{0pt}
    \subcaptionbox{Grid-level outputs over 4 days. Similar weather conditions occur on day 2 and 3, but with different capacities. \label{fig:grid_output}}{
        \begin{minipage}[t]{\textwidth}
            \vspace{0pt}
            \includegraphics[width=\linewidth]{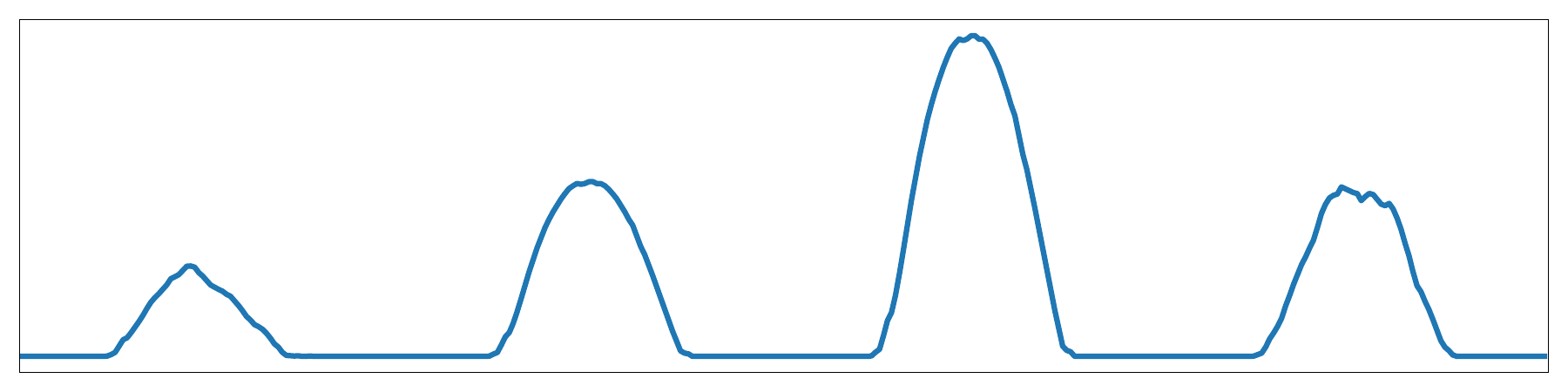}\par\vfill
            \includegraphics[width=\linewidth]{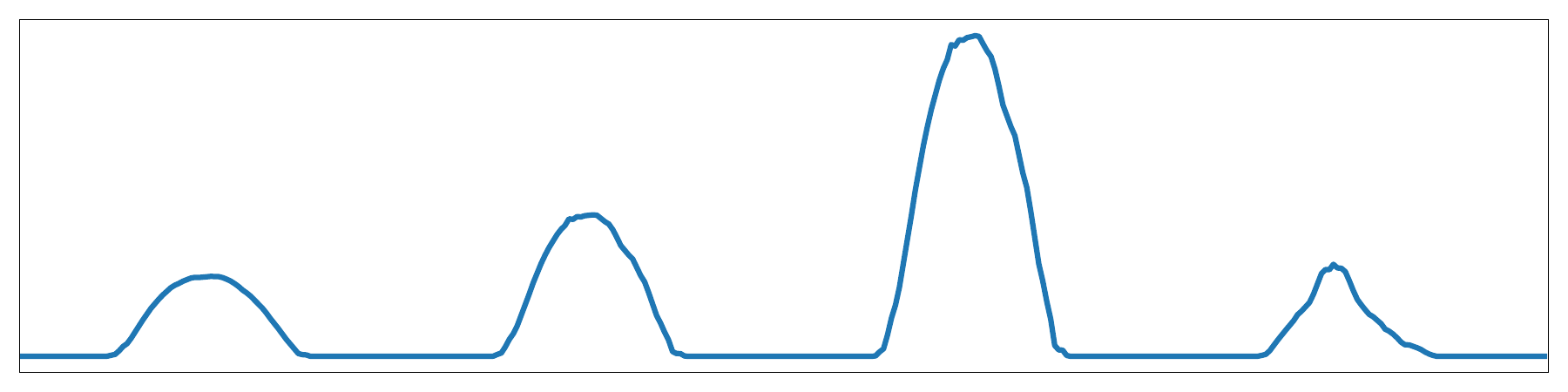}\par\vfill
            \includegraphics[width=\linewidth]{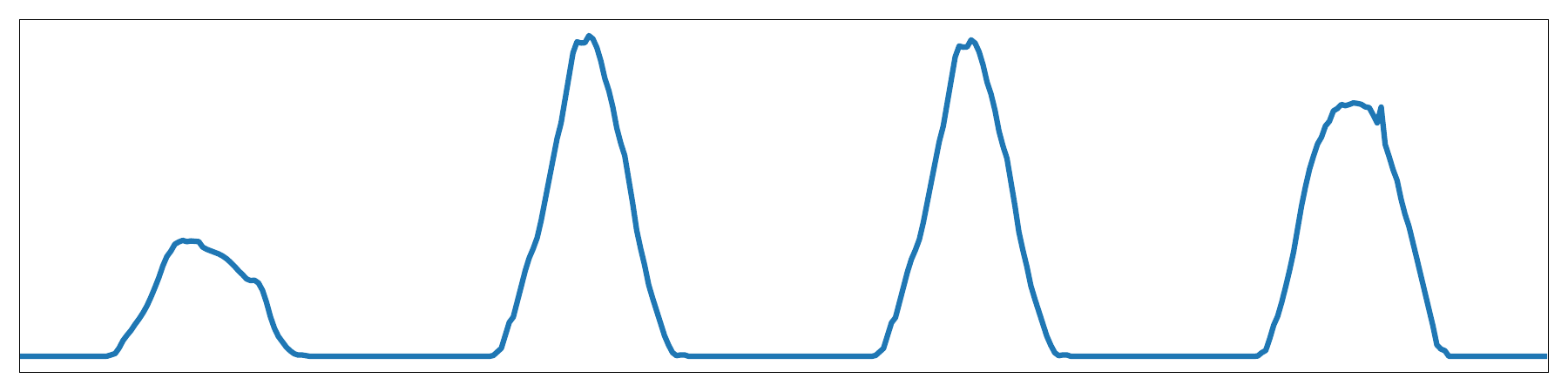}\par\vfill
            \vspace{.2cm}
            \includegraphics[width=\linewidth]{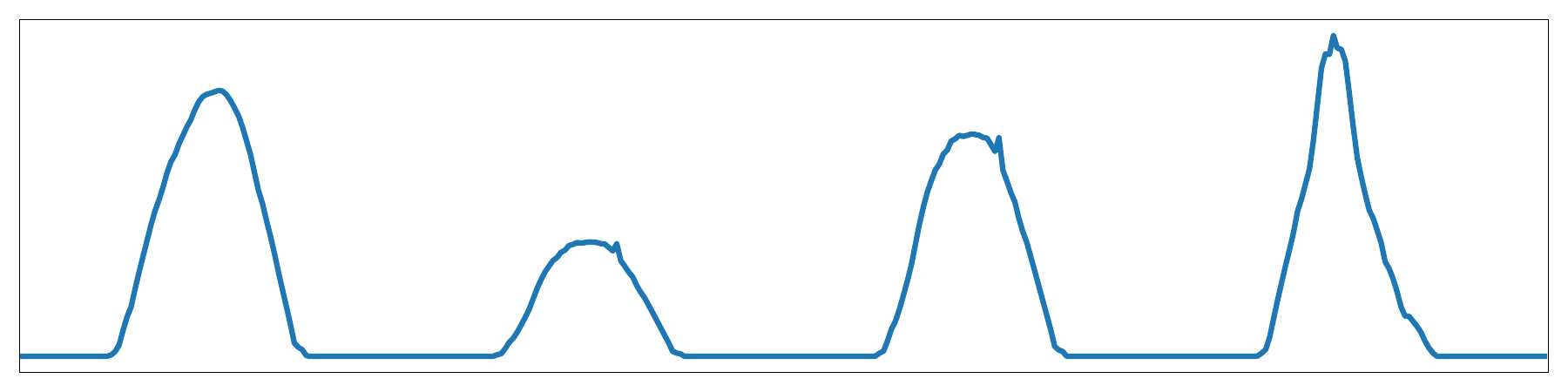}\vspace{-.15cm}
        \end{minipage}
    }
\end{minipage}%
\hspace{.3cm}
\begin{minipage}[t][][t]{0.42\linewidth}
    \vspace*{0pt}
    \subcaptionbox{Total capacities stacked by 6 grid-level capacities.\label{fig:capacity_bar}}{
        \includegraphics[width=\linewidth]{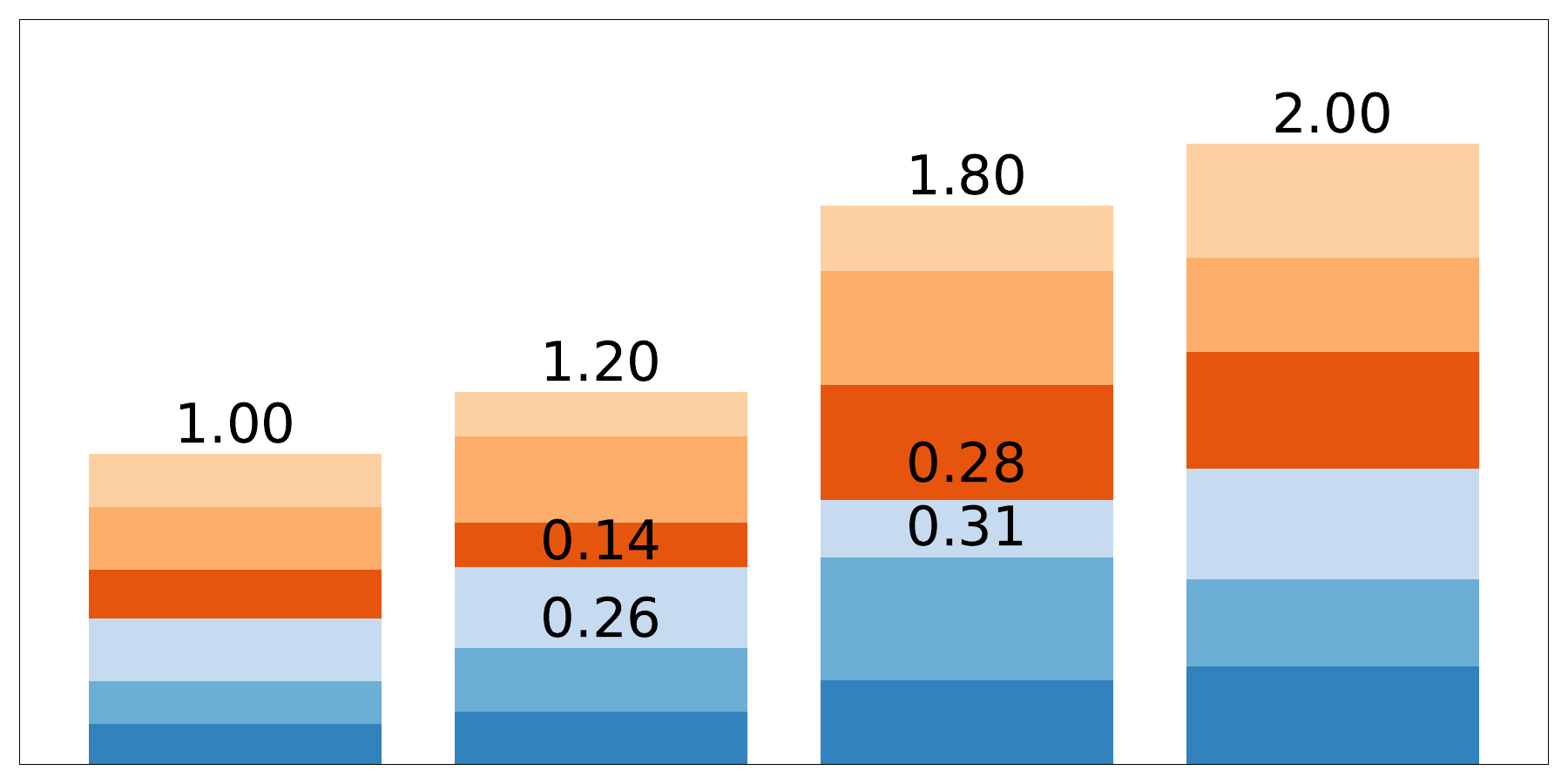}\vspace{-.1cm}
    }
    
    \subcaptionbox{True total outputs (blue) vs. scaling estimation (orange). \label{fig:total_output}}{
        \includegraphics[width=\linewidth]{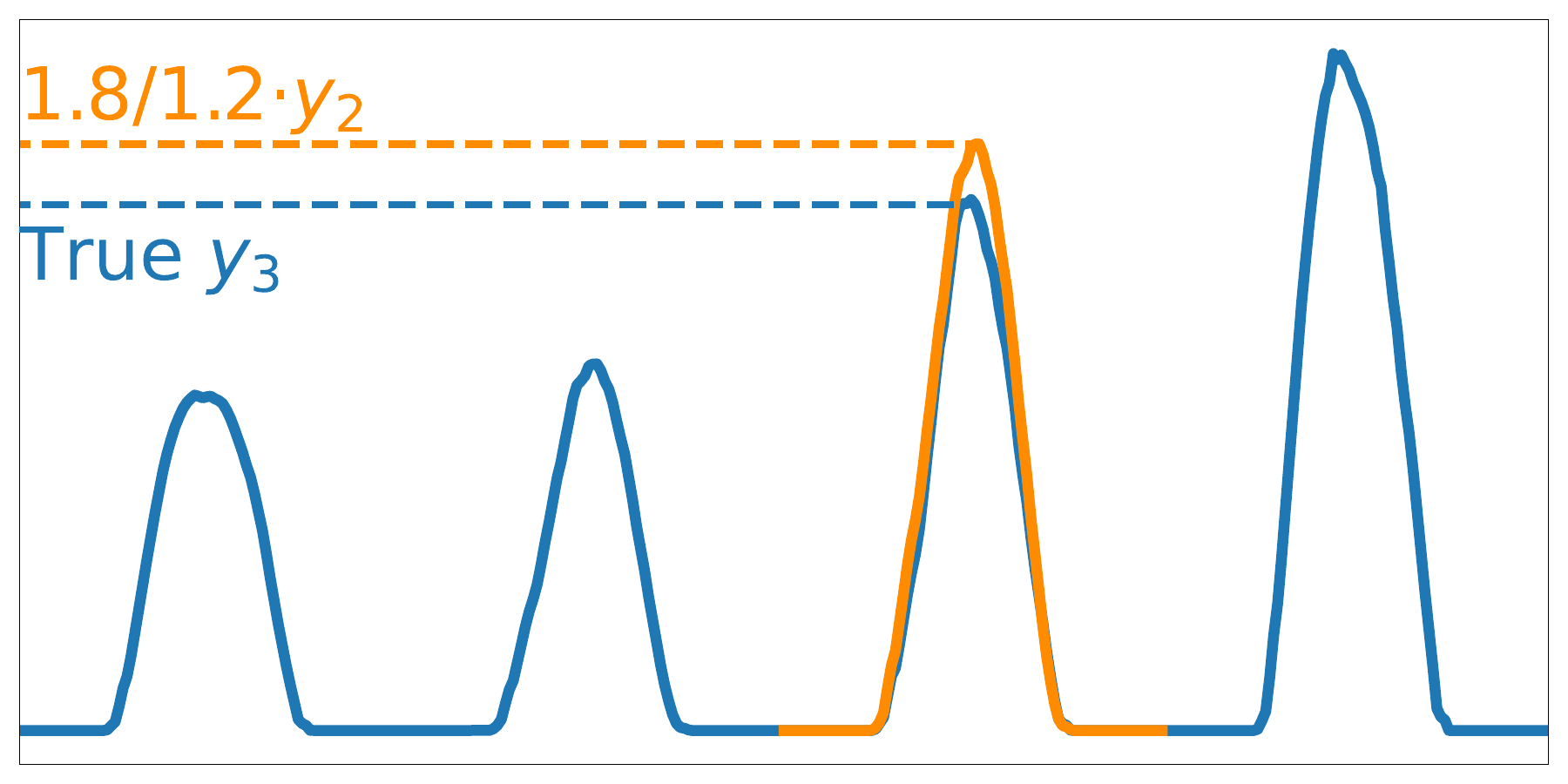}\vspace{-.12cm}
    }
\end{minipage}
\vspace{-.6cm}
\caption{Varying grid-level capacities proportions (b) across time make grid-level modeling necessary: relying solely on total capacity (orange curve in (c)) results in large errors (difference between orange and blue dashed lines in (c)), even under similar weather conditions between day 2 and 3 in (a).}
\vspace{-.6cm}
\label{fig:agg}
\end{figure}

Grid-level capacity determines each grid's contribution to the total output. As shown in Figure~\ref{fig:agg} (c), ignoring grid-level capacity dynamics leads to estimation errors (gap between dashed lines), highlighting the need for grid-level capacity modeling (see Observation \ref{obs:flattening}).

\begin{observation}\label{obs:flattening}
\textbf{Grid-Level Capacities as True Multipliers}. Consider $\bm{x'}$ that aggregates $\bm{x}$ from $T\times D\times K$ into $T\times(D\cdot K)$, with an aggregation function $f_\text{agg}(\cdot)$ that models the total output at time $t-1$, as defined in Equation \eqref{eq:fagg}. If the capacity of grid $i$ changes from $c_{t-1,i}$ at $t-1$, to $c_{t,i}$ at $t$, merely adjusting the output by the total capacity ratio $r=C_t/C_{t-1}$ does not  generalize the prediction at time $t$, as shown in Equation \eqref{eq:fagg2}:
\begin{align}
f_{\text{agg}}(\bm{x'}) :=&\sum_{i=1}^K c_{t-1,i} \cdot f(\bm{x}_i) \label{eq:fagg}
\\ \not\Rightarrow
r f_{\text{agg}}(\bm{x'})=&\sum_{i=1}^K c_{t,i} \cdot f(\bm{x}_i) ,\ \ r:=C_t/C_{t-1}.\label{eq:fagg2}
\end{align}
\end{observation}
In essence, grid-level capacities should be integrated into the model, either explicitly or latently, as multiplier indicators.

%% file: 3.method.tex
\section{Problem Formulation}
\label{sec:Problem}







\sloppy
Consider a geographical area divided into \( K \) grids. For each time step \( t \), let \( \boldsymbol{x}_{t,i} \in \mathbb{R}^D \) represent the input features of grid $i$, including predictive weather forecasts and auxiliary covariates. Thus the input for the entire area is \( \boldsymbol{x}_t := [\boldsymbol{x}_{t,1}, \boldsymbol{x}_{t,2}, \dots, \boldsymbol{x}_{t,K}] \in \mathbb{R}^{K \times D} \). The corresponding PV power output of the entire area at time \( t \) is denoted by a single value \( Y_t \in \mathbb{R} \). Given $T$ time steps, the inputs area structured as a tensor of dimensions $T\times K\times D$, and the outputs are structured as a vector of dimension $T$.

Motivated by Observation \ref{obs:multiplier} and \ref{obs:flattening}, we introduce grid-level capacity and per-unit-capacity output function into our model. To address the challenge that the real-world unit output varies across PV panels, we incorporate heterogeneity into the grid-level capacity, defined as \textit{generalized-capacity}. This generalized-capacity of grid $i$ is not a precise measurement of the capacity, but rather a variable that encapsulates multiple influencing factors, including capacity, efficiency conversion decay, and different characteristics in PV panel. This allows all heterogeneity to be captured within the generalized-capacity and the unit output function is able to maintain a consistent mapping relationship. We introduce the concept of generalized capacity to represent the effective contribution of each site to the overall output, decoupling capacity from unit output function.

Formally, let \( \boldsymbol{c}_t := [c_{t,1}, c_{t,2}, \dots, c_{t,K}] \in \mathbb{R}^{K} \)  be the generalized-capacity at time $t$, where \( c_{t,i} \in \mathbb{R} \) is the composite value of grid $i$.  We define $f:\mathbb{R}^D \times \Theta \rightarrow \mathbb{R}$ as the per-unit-capacity output function with learnable parameters $\bm \theta\in \Theta$, which maps the input features $\boldsymbol{x}_{t,i}$ to the predicted per-unit-capacity output. This output approximates the unobserved $y_{t,i}$ such that $Y_t = \sum_{i=1}^{K} c_{t,i} y_{t,i}$. Consequently, the predicted output $\hat Y_t$ for the entire area at time $t$ is given by:
\begin{equation}
    \hat Y_t = \sum_{i=1}^{K} c_{t,i} f(\boldsymbol{x}_{t,i};\boldsymbol{\theta}).
\end{equation}
 

Our objective is to learn both the pre-unit-capacity output predictor $f(\cdot; \boldsymbol{\theta})$ and generalized-capacity $\boldsymbol{c}_t$ throughout the whole $T$-step period by the following minimization problem:
\begin{equation}
\min_{\boldsymbol{\theta}, \{\boldsymbol{c}_{t}\}_{t=1}^T} \sum_{t=1}^{T} \left( Y_t - \sum_{i=1}^{K} c_{t,i} f(\boldsymbol{x}_{t,i}; \boldsymbol{\theta}) \right)^2 \quad \text{s.t.} \quad \boldsymbol{c}_{t} \in \Omega_t,\forall t\in T,
\label{eq:obj}
\end{equation}
where $\Omega_t$ is a constraints set of $\boldsymbol{c}_t$.
All capacities are non-negative, and they sum up to the known variable of total capacity $C_t$. Based on observations in practice, we assume that the capacity changes gradually over time, controlled by the cap value $\epsilon$ for the gap between two adjacent time steps. Specifically, $\Omega_t$ is defined as:
\begin{equation}
\Omega_t := \left\{\, \bm{c}_t \in \mathbb{R}^K \;\middle|\;
\begin{aligned}
&\sum_{i=1}^{K} c_{t,i} = C_t, \, c_{t,i} \geq 0 \quad \forall i &\text{(a)}\\
&\|\boldsymbol{c}_{t} - \boldsymbol{c}_{t-1}\|_2 \leq \epsilon, \quad t \geq 2 &\text{(b)}
\end{aligned}
\,\right\}.
\label{eq:conditions}
\end{equation}
Here, conditions (a) and (b) represent the numerical and temporal constraint, respectively. $\|\cdot\|_2$ denotes the $L_2$-norm. For $t=1$, constraint (b) does not apply.

Notational conventions are as follows: unless otherwise specified, uppercase letters denote aggregated variables (single value per time step), lowercase letters denote unit variables (values per grid and time step), boldface symbols represent vectors, non-boldface symbols denote scalars, superscripts indicate iteration count, and subscripts denote time steps and grid indices.


\section{SolarBoost}
\label{sec:SolarBoost}
We propose SolarBoost, a novel framework for predicting the output power of DPV systems and efficiently solving the optimization problem. 

\begin{figure}[t]
    \centering
\includegraphics[width=1\linewidth]{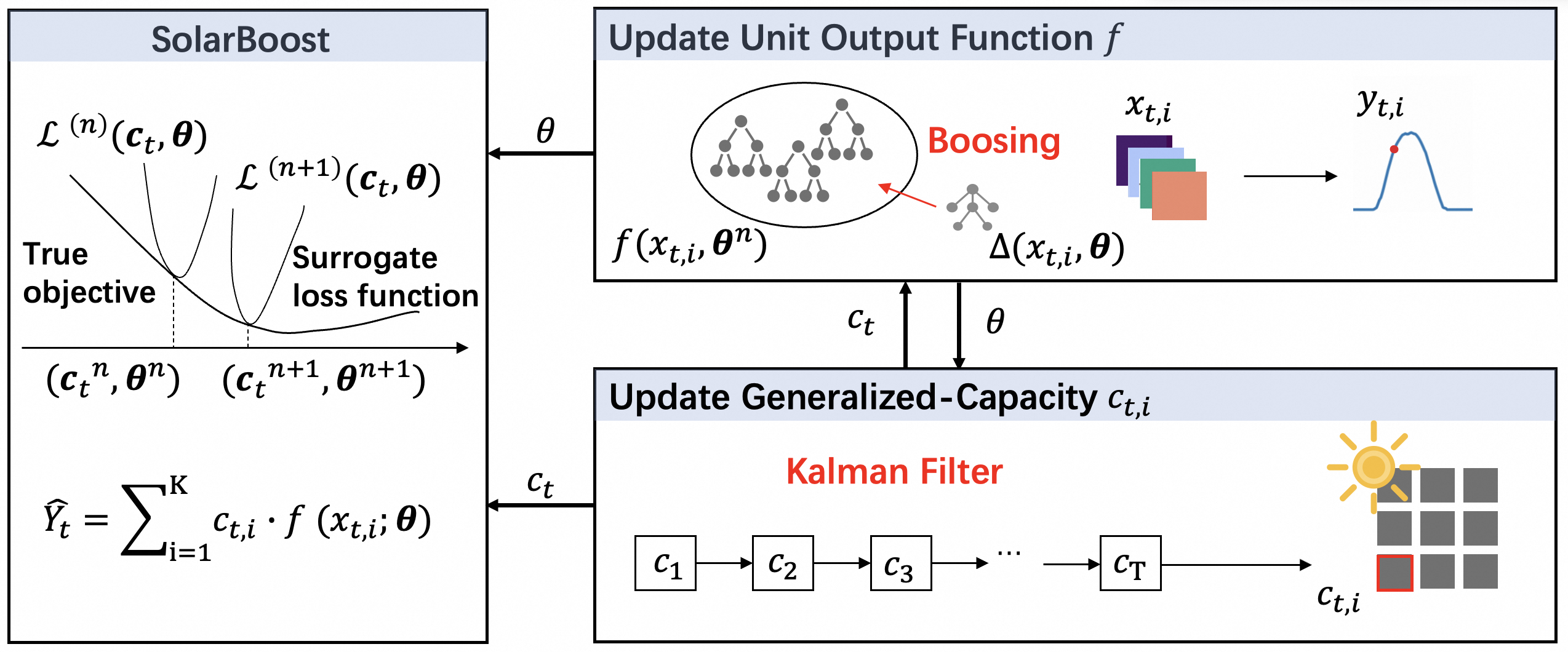}
    \vspace{-.5cm}
    \caption{The framework of SolarBoost, which uses a surrogate loss function to approximate the true objective function and alternating updates of unit output predictor and generalized-capacity by  boosting and Kalman filter, respectively.
}
    \label{fig:SolarBoost}
    \vspace{-.6cm}
\end{figure}

Since the original optimization  in Equation~\eqref{eq:obj} with nested summations inside the $\ell_2$-norm is intractable, we draw inspiration from the iterative process over a surrogate objective function used in the Majorization-Minimization (MM) method \cite{hunter2004tutorial}. Specifically, we first design our own upper bound and then employ an iterative minimization procedure for estimating the unit output predictor $f$ and the generalized-capacities $\boldsymbol{c}_{t}$:  $f$ is updated using a boosting method, while $\boldsymbol{c}_{t}$ are estimated using a Kalman filter.
This alternating minimization ensures a monotonic decrease in the true objective of Equation~\eqref{eq:obj} while maintaining computational tractability. The whole process is illustrated in Figure~\ref{fig:SolarBoost}.

Section~\ref{subsec:surrogate_loss}  introduces our surrogate loss function. In Sections ~\ref{subsec:unit_output_predictor} and ~\ref{subsec:capacity-estimate}, we detail the procedures for alternately updating the unit output predictor and the generalized capacities, respectively.

\subsection{Surrogate Loss Function}
\label{subsec:surrogate_loss}
We construct a tight upper bound of the true objective function, and then minimizing this function. We rewrite the true objective function in Equation~\eqref{eq:obj} for time step $t$ as:
\begin{align}
\begin{split}
\mathcal{O}_t(\boldsymbol{\theta},\boldsymbol{c}_{t}):=\left(Y_t - \sum_{i=1}^{K} c_{t,i} f(\boldsymbol{x}_{t,i}; \boldsymbol{\theta}) \right)^2 
    = \| \boldsymbol{c}_t^{T} \boldsymbol{r}_t \|_2^2,
\label{upperbound_t}
\end{split}
\end{align}
where $\boldsymbol{r}_t=[r_{t,1}, r_{t,2}, \dots, r_{t,K}] \in \mathbb{R}^{K}$ is the fitting residuals, with $r_{t,i} = y_{t,i} - f(\boldsymbol{x}_{t,i},\boldsymbol{\theta})$ denoting the residual for grid $i$.
Since $\| \cdot \|_2^2$ is convex, we can derive a tight upper bound for $\mathcal{O}_t$ via lemma \ref{lemma1}.
\begin{lemma}
\label{lemma1}
A convex function $\kappa(\cdot)$ satisfies the inequality:
$$\kappa(\sum_i \alpha_it_i) \leq \sum_i\alpha_i\kappa(t_i),$$
where $\alpha_i \geq 0$ and $\sum_i \alpha_i = 1$ for any collection of points $t_i$ and corresponding multipliers $\alpha_i$. For a linear function $\bm{z}^T \bm{\beta}$, the following inequality holds:
$$\kappa(\boldsymbol{z}^T \boldsymbol{\beta}) \leq \sum_i \alpha_i \kappa[\frac{z_i}{\alpha_i}(\beta_i-\beta_i^{(n)})+\boldsymbol{z}^T\boldsymbol{\beta} ^{(n)} ].$$
\end{lemma}
By setting $\kappa(\cdot) = (\cdot)^2$, $\alpha_i = 1/K$ for all $i$, $\boldsymbol{z} = \boldsymbol{c}_t$ and $\boldsymbol{\beta} = \boldsymbol{r}_t$, we obtain the following upper bound for $\mathcal{O}_t$ at iteration $n+1$:
\begin{align}
\begin{split}
    \mathcal{L}_t^{(n+1)}(\boldsymbol{\theta}, \boldsymbol{c}_{t}) =& \sum_{i=1}^{K} \frac{1}{K} [ K c_{t,i} \left(r_{t,i} - r^{(n)}_{t,i}\right) + \boldsymbol{c}_t^{T} \boldsymbol{r}_t^{(n)}) ]^2 \\
    =& \sum_{i=1}^{K} \frac{1}{K} [ -K c_{t,i} \left( f(\boldsymbol{x}_{t,i}; \boldsymbol{\theta}) - f(\boldsymbol{x}_{t,i}; \boldsymbol{\theta}^{(n)})\right) + \boldsymbol{c}_t^{T} \boldsymbol{r}_t^{(n)}]^2.
\label{eq:upperbound_t_n}
\end{split}
\end{align}
In contrast to the formulation in (\ref{eq:obj}), where the weighted summation of the function 
$f(\cdot;\boldsymbol{\theta})$ across different regions results in feature entanglement, our surrogate function $\mathcal{L}_t^{(n+1)}(\boldsymbol{\theta}, \boldsymbol{c}_{t})$ utilizes region-specific decoupling. Specifically, each regional instance of $f(\boldsymbol{x}_{t,i};\boldsymbol{\theta})$ maintains an independent fitting objective under $\mathcal{L}_t^{(n+1)}(\boldsymbol{\theta}, \boldsymbol{c}_{t})$. This approach allows us to eliminate cross-region interference during optimization, facilitating the simultaneous training of functions across different areas.

We now address the solution to solving equation (\ref{eq:obj}) by iteratively minimizing the surrogate function $\mathcal{L}_t^{(n+1)}(\boldsymbol{\theta}, \boldsymbol{c}_{t})$. Since \( f(\cdot;\boldsymbol{\theta}) \) and the capacity \(\boldsymbol{c}_{t}\) remain intertwined, we will optimize them alternately in the following subsections.

\subsection{Update Unit Output Function $f$ via Boosting}
\label{subsec:unit_output_predictor}
We aim to update \( f(\boldsymbol{x}_{t,i}; \boldsymbol{\theta}) \) by minimizing the surrogate loss function \eqref{eq:upperbound_t_n} while keeping the generalized-capacities \(\{\boldsymbol{c}_t\}_{t=1}^T\) fixed. 

In the formulation \eqref{eq:upperbound_t_n}, the 
$(n+1)$-th iteration requires minimizing the surrogate function by updating the parameter $\boldsymbol{\theta}$. However, retraining \( f(\boldsymbol{x}_{t,i}; \boldsymbol{\theta}) \) from scratch in each iteration is computationally expensive and may lead to overfitting, especially when dealing with large datasets or complex models. 

To address these issues, we use a residual boosting reparameterization approach.
We observe that the surrogate loss $\mathcal{L}$ fundamentally depends on predictive residuals as defined by
\begin{equation}
\Delta_t(\boldsymbol{x}_{t,i}) = f(\boldsymbol{x}_{t,i}; \boldsymbol{\theta}) - f(\boldsymbol{x}_{t,i}; \boldsymbol{\theta}^{(n)}).
\end{equation}
With this observation, we reformulate the optimization process to focus on learning incremental corrections 
$\Delta_t$ rather than conducting full model retraining. 
By substituting \(\Delta_t(\boldsymbol{x}_{t,i})\) into \eqref{eq:upperbound_t_n}, we reformulate the upper bound for the objective function at time step \( t \) in the \((n+1)\)-th iteration:
\[
\mathcal{L}_t^{(n+1)}(\boldsymbol{\theta}, \boldsymbol{c}_{t}) = \sum_{i=1}^K \frac{1}{K} \left[ -K c_{t,i} \Delta_t(\boldsymbol{x}_{t,i}) + \boldsymbol{c}_t^\top \boldsymbol{r}_t^{(n)} \right]^2.  
\] 
In the $(n+1)$-th iteration, we identify an ensemble component $\Delta_t^{(n+1)}(\boldsymbol{x}_{t,i})$
that minimizes the surrogate loss $\mathcal{L}$.
The updated prediction function is then expressed as \( f(\boldsymbol{x}_{t,i}; \boldsymbol{\theta}^{(n+1)})=\sum_{j=1}^{n+1} \eta\Delta_t^{(j)}(\boldsymbol{x}_{t,i})\), where this sum represents the cumulative prediction from
$n+1$ weak learners, and $\eta$ denotes the learning rate.

In this paper, we implement the described boosting framework using Gradient Boosted Decision Trees (GBDT), taking advantage of advanced state-of-the-art (SOTA) implementations such as LightGBM and XGBoost. These tools efficiently handle the computations needed for training, leveraging provided gradients and Hessians of the loss function with respect to
$\Delta_t(\boldsymbol{x}_{t,i}; \boldsymbol{\theta})$. Derived from our surrogate loss function, these formulations are as follows:
\begin{align}
\text{grad}^{(n+1)}_{t,i} &= -2c_{t,i}^{(n)} \left( \boldsymbol{c}_t^\top \boldsymbol{r}_t - \sum_{i=1}^K c_{t,i}^{(n)} \Delta_t(\boldsymbol{x}_{t,i}; \boldsymbol{\theta}) \right)\label{grad},\\
\text{hess}^{(n+1)}_{t,i} &= 2 \left( c_{t,i}^{(n)} \right)^2.
\label{hess}
\end{align}

These formulations show that our surrogate loss function assigns higher weights to grids with larger generalized capacities, directing the model's focus to the most important areas for improved predictive performance.


\subsection{Update Generalized-Capacity $c_{i,t}$}
\label{subsec:capacity-estimate}
We aim to update the generalized-capacity series $\{\boldsymbol{c}_t\}$ by minimizing the surrogate function~\eqref{eq:upperbound_t_n} subject to the constraints presented in (\ref{eq:conditions}).
Before proceeding, we first express the surrogate loss function \eqref{eq:upperbound_t_n} in matrix form as follows:
\begin{equation}
\mathcal{L}_t^{(n+1)} (\boldsymbol{\theta}, c_{t,i})
= \left\| \boldsymbol{\Sigma}_t^{\frac{1}{2}} \boldsymbol{c}_t - Y_t \boldsymbol{\Sigma}_t^{-\frac{1}{2}}(\boldsymbol{\Delta}_t + \boldsymbol{q}_t) \right\|_2^2,
\label{upperbound_t_n_matrix}
\end{equation}
where 
\begin{align}
    \boldsymbol{q}_t =& \left[ f(\boldsymbol{x}_{t,1}; \boldsymbol{\theta}^{(n)}), \ldots, f(\boldsymbol{x}_{t,K}; \boldsymbol{\theta}^{(n)}) \right]^T\\
    \boldsymbol{\Delta}_t =&[\Delta_t(\boldsymbol{x}_{t,1}; \boldsymbol{\theta}), \ldots, \Delta_t(\boldsymbol{x}_{t,K}; \boldsymbol{\theta})]^T\\
    \boldsymbol{\Sigma}_t =& K \boldsymbol{Q}_t + \boldsymbol{q}_t \boldsymbol{q}_t^T + \boldsymbol{q}_t \boldsymbol{\Delta}_t^T + \boldsymbol{\Delta}_t \boldsymbol{q}_t^T\\
    \boldsymbol{Q_t} =& \mathrm{diag}([\Delta_t (\boldsymbol{x}_{t,1}; \boldsymbol{\theta})^2, \ldots, \Delta_t (\boldsymbol{x}_{t,K}; \boldsymbol{\theta})^2]).
\end{align}
Here $\boldsymbol{q}_t$ and $\boldsymbol{\Delta}_t$ 
represent predictions from the first $n$ learners and the $(n+1)$-th learner for all grids at a given time step $t$, respectively.

Optimizing $\boldsymbol{c}_t$ involves solving a least squares problem under a set of constraints. 
To simplify this, we convert the constrained optimization problem into an unconstrained one by adding a Lagrangian penalty term:
\begin{equation}
\min_{\{\boldsymbol{c_t}\}}\sum_{t=1}^T\left\| \boldsymbol{\Sigma}_t^{\frac{1}{2}} \boldsymbol{c}_t - Y_t \boldsymbol{\Sigma}_t^{-\frac{1}{2}} (\boldsymbol{\Delta}_t + \boldsymbol{q}_t) \right\|_2^2 + \lambda \| \boldsymbol{c}_{t} - \boldsymbol{c}_{t-1} \|_2^2, 
\label{opt_capacity}
\end{equation}
where $\lambda$ is a hyperparameter that balances prediction accuracy and temporal continuity of the generalized-capacity.
The above formulation only addresses constraint (b) in $\Omega_t$. The discussion for handling constraint (a)will be in Section~\ref{subsec:overall_framework}.

Simultaneous optimization of all $\{\boldsymbol{c}_{t}\}_{t=1}^T$ involves $T \times K$ parameters, leading to a prohibitive computational complexity of $O(T^3 K^3)$. 
To mitigate this, 
we propose updating $\boldsymbol{c}_t$ using a dynamic Kalman filtering procedure.
The Kalman filter is designed to estimate hidden states in a linear dynamic system. In our paper, we develop a customized Kalman filtering framework to estimate the hidden state sequence 
$\{\boldsymbol{c}_t\}_{t=1}^T$
within a linear dynamical system. Formally, the state-space model is constructed as follows:
\begin{itemize}[leftmargin=1em]
\item \textbf{State Transition Model}: The temporal evolution of generalized capacities follows
$\boldsymbol{c}_t=\boldsymbol{c}_{t-1}+\boldsymbol{w}_t$, where $\boldsymbol{w_t}\sim \mathcal{N}(0,\gamma^{-1}\boldsymbol{I})$.

\item \textbf{Observation Model}: Measurements are defined through the following mapping:
\begin{equation}
\boldsymbol{\eta}_t = \boldsymbol{\Sigma}_t^{1/2} \boldsymbol{c}_t + \boldsymbol{v}_t, \quad \boldsymbol{v}_t \sim \mathcal{N}(\boldsymbol{0}, \boldsymbol{I})
\end{equation}
where $\boldsymbol{\eta}_t \triangleq Y_t \boldsymbol{\Sigma}_t^{-1/2}(\boldsymbol{\Delta}_t + \boldsymbol{q}_t)$ represents preprocessed observations, and the time-varying sensing matrix $\boldsymbol{\Sigma}_t^{1/2}$ encodes spatial correlations derived from learner ensembles.

\item \textbf{Time-Varying Adaptation}: Unlike conventional Kalman filters with static sensing matrices, our formulation dynamically updates $\boldsymbol{\Sigma}_t^{1/2}$ based on the incremental learner's predictions $\boldsymbol{\Delta}_t$ and historical outputs $\boldsymbol{q}_t$. This adaptation mechanism enables context-aware adjustments to distribution shifts.

\end{itemize}
The Kalman recursion provides sequential updates with $\mathcal{O}(TK^3)$ complexity—a polynomial reduction from the original cubic scaling. This enables real-time estimation even for large-scale systems. Additional computational experiments and analysis can be found in Appendix~\ref{app:Computational}.

\subsection{Overall Framework}
\label{subsec:overall_framework}
Our proposed method, SolarBoost, is illustrated in Figure \ref{fig:SolarBoost} and alternates updates between  the unit output predictor \( f(\boldsymbol{x}_{t,i}, \boldsymbol{\theta}) \) and the generalized-capacity \( \{\boldsymbol{c}_t\}_{t=1}^T \), to minimize a surrogate loss function. 
This surrogate loss function decouples the parameters $\boldsymbol{\theta}$ from the set $\{\boldsymbol{c}_{t}\}_{t=1}^T$, thereby facilitating a more tractable minimization process. 
The Unit Output Predictor employs a gradient boosting trees framework to model the pre-unit-capacity output across \( K \) spatial grids and \( T \) time steps. In each iteration, a new tree is trained to update the predictions.
The generalized-capacity is updated using a dynamic Kalman filtering procedure. 

In each iteration \( n \), we initialize \( \boldsymbol{c}_1 \) with the average capacity of each grid from all previous time steps (\( c_{1,i} = \frac{C_t}{K} \)) to ensure numerical stability. The generalized-capacity time series is divided into subsequences of length $s$, keeping capacity constant within each subsequence and therefore removing constraint (a) in Equation~\eqref{eq:conditions}. 
To predict future power output \( Y_{T+h} \) for horizon \( h \), we input the forthcoming features \( \{\boldsymbol{x}_{T+h}\} \) into the unit output predictor and use the latest generalized-capacity \( \{\boldsymbol{c}_{T}\} \) to calculate the total power output.
We summarize our proposed SolarBoost
in Algorithm~\ref{alg}.


\begin{algorithm}
\caption{SolarBoost Algorithm}
\label{alg}
\KwInput{Dataset $\{\boldsymbol{x}_t, Y_t\}_{t=1}^T$, number of iterations $N$}
\KwOutput{Aggregate output predictions $\{ \hat{Y}_t \}_{t=1}^T$}
\For{iteration $n \gets 1$ \KwTo $N$}{
    \textbf{Step 1: Train Unit Output Predictor}\\
    Train a regression tree $\Delta(\boldsymbol{x}_{t,i}; \boldsymbol{\theta})$ using ~\eqref{grad} and ~\eqref{hess}\\
    
    \textbf{Step 2: Generalized-Capacity Estimation}\\
    Initialize prior state $\boldsymbol{c}_1^{(n+1)}$\\
    Update $\{\boldsymbol{c}^{(n+1)}_t\}_{t=2}^T$ using Kalman filter\\
        
}

\textbf{Step 3: Compute Aggregate Predictions}\\
$\hat{Y}_t \gets \sum_{i=1}^K c_{t,i}^{(N)} f(\boldsymbol{x}_{t,i}; \boldsymbol{\theta}^{(N)})$
\end{algorithm}

%% file: 4.theoretical_analysis.tex
\section{Theoretical Analysis}
\label{sec:thm}
\begin{theorem}[Superiority of Grid-Level Modeling]
\label{thm:sup}
    Let $\{c_{t,i}\}_{i=1}^K$ denote $K$ grid-level capacities at time $t$, and assume that for all $t \geq 2$ and $i$, the capacity changes satisfy $|c_{t,i} - c_{t-1,i}| \leq \epsilon$. Let $Y$ be a nonnegative target variable satisfying $|Y| \leq M$ almost surely. There exists a sample size $N$ such that grid-level modeling with $c_{t,i}$  achieves a tighter (lower) upper bound than aggregate modeling, provided that
    \[
        N \gtrsim    \frac{K\sigma_c^2M^2}{r^2K^2(1+K)^2\epsilon^2 M^2-C_t^2\sigma_f^2},
    \]\vspace{-.3cm}
\end{theorem}
where $\sigma^2_c$ and $\sigma_f^2$ are related to the variance of $\{c_{t,i}\}_{i=1}^K$ and $f(\cdot)$, respectively. 
Aggregate modeling introduces significant bias though  maintains low variance, whereas grid-level modeling yields unbiased estimates with a larger variance. Theorem~\ref{thm:sup} implies that  with a sufficiently large sample size $N$ for each $c_{t,i}$, grid-level modeling outperforms aggregate modeling. Such a sample size is generally attainable, and our ablation study empirically confirms the advantages of grid-level modeling. In addition, a larger capacity change $\epsilon$  would better suggest to adopt grid-level modeling.

\begin{theorem}{Allowance for $c$ with high variance}.
\label{thm:var}
Let $f'(\bm x_t)$ $ = \sum_i c_{t,i} f(x_{t,i})$ be a well-specified function where $\bm c_t = (c_{t,1},c_{t,2},\ldots,c_{t,K})$ are the parameters to be estimated. Then for any $\bm x_t$, under a fixed learning error $\mathbb{E}[\hat f'(\bm x_t)-f'(\bm x_t)]^2$,
$$Var(\bm{c_t})\propto\frac{1}{\sigma_x^2},$$
where $\sigma_x^2$ denotes the variance of the inputs $\{x_{t,i}\}_{i=1}^K$.
\end{theorem}

The variance of $x$ is inversely proportional to the variance of $c$. In other words, when the fluctuations in $x$ are small and the inputs at different grid points are highly correlated, we can tolerate less accurate estimations of $c$, yet still achieve the same predictive accuracy for $Y$ as we would if $x$ were uncorrelated and $c$ were accurately estimated.









%% file: 5.experiment.tex
\section{Experiment}
\subsection{Setup}
\subsubsection{Baselines}

We consider two core baseline methods that employ common multigrid data handling paradigms as described in Section ~\ref{subsec:grid_capacity}, and are integrated with a standard regression model. 

\begin{itemize}[leftmargin = .2cm]
    \item \textbf{AverageGrid}: calculates the mean of features across all grids in the entire region, using this $D$-dimensional average vector as the input for training a regression model to predict the output. 
    \item \textbf{FlattenGrid}: treats the features of the entire region as a tensor, flattens it into a $K \cdot D$-dimensional vector, and uses the resulting vector as input to train a regression model.
\end{itemize}
In synthetic data experiments, we also include an \textbf{IdealFit} baseline, representing an idealized scenario where the regression model directly fits the correctly-specified function with data of all time steps.where
In real-world data experiments, we incorporate two additional state-of-the-art methods to provide a more comprehensive evaluation: \textbf{CNN-LSTM} \cite{NIPS2015_07563a3f}(a neural network-based approach) and \textbf{SARIMAX}(a statistical method). We apply Averaging and Flattening multigrid data handling paradigms to SARIMAX, referring to these combinations as Ave-SARIMAX and Flat-SARIMAX respectively. 
Direct comparison with existing DPV forecasting methods mentioned in Appendix~\ref{app:related_work} is limited by the availability of reproducible implementations. The neural network and statistical baselines provide baseline diversity.

\subsubsection{Implementation}
Due to space constraints, detailed implementation settings are provided in Appendix~\ref{app:implementation}.

\subsubsection{Evaluation Metric}
For evaluation, we employ the root mean square error (RMSE), a standard metric for assessing the accuracy of PV forecasting. RMSE is computed as follows:
$\text{RMSE}=\sqrt{\frac{1}{T}\sum^{T}_{t=1}(Y_t-\hat{Y}_t)^2}$,
where $T$ denotes the time range of test set, $Y_t$ is the ground truth at time $t$, and $\hat{Y}_t$ is the predicted value at time $t$.

\subsection{Synthetic Data}
To evaluate the model's ability to capture temporal shifts and make accurate predictions, we designed toy dataset experiments to answer the following questions:
\begin{itemize}[leftmargin = .2cm]
    \item \textbf{Q1}: Can our model capture the information of varying generalized capacity across different grids and their temporal evolution?
    \item \textbf{Q2}: Can our model predict the unit output accurately?
    \item \textbf{Q3}: Can our model predict the aggregated output accurately?
\end{itemize}

We begin by introducing the synthetic datasets in Section ~\ref{subsubsec:toy_data}, followed by answers to the three questions in Sections ~\ref{subsubsec:Q1}--\ref{subsubsec:Q3}.

\subsubsection{Dataset Description}
\label{subsubsec:toy_data}
\sloppy
Each synthetic dataset comprises three essential components: an input tensor $\{\boldsymbol{x}_{t,i,d}\}$, a generalized capacity matrix $\{c_{t,i}\}$, and a response variable vector $\{Y_{t}\}$. Here, $\{\boldsymbol{x}_{t,i,d}\}$ represents the input features for time step $t$, grid $i$, and feature dimension $d$; the generalized capacity matrix $\{c_{t,i}\}$ scales the influence of each grid $i$ at time $t$ on the final output. To evaluate the models' ability to capture generalized capacity effects, we generate capacity matrices using two methods: the AR(1) model (\textit{AR} dataset, independent capacity dimensions) and the Kalman filter (\textit{Kalman} dataset, correlated capacity dimensions).
An intermediate response $y_{t,i}$ for each time step $t$ and grid $i$ is calculated using a transformation of the input features:
\begin{equation}
    y_{t,i} = sin(x_{t,i,1}) + x_{t,i,2} + x_{t,i,3}^2.
\label{eq:function}
\end{equation}
Finally, the output variable ${Y_t}$ is obtained by scaling and aggregating the intermediate responses with the corresponding generalized capacities: $Y_{t} = \sum^K_{i=1}c_{t,i} \cdot y_{t,i}.$ 

The dataset comprises 28800 samples, with indices representing time steps $t \in \{1, \dotsc, 28800\}$, grids $i \in \{1,\dotsc 15\}$, and input features $d \in \{1,2,3\}$. We use the initial 26880 samples for training and the final 1920 samples for testing. See Appendix~\ref{app:toy_data} for details on the data generation process.

\subsubsection{Generalized Capacity Forecasting Performance}
\label{subsubsec:Q1}


Among the baselines, only AverageGrid attempts to forecast grid-level capacities, albeit with a simplifying assumption of uniform generalized-capacity $(C_t/K)$ across all grids at each time step. To quantitatively assess generalized capacity forecasting, we calculate the RMSE between true and predicted capacities. As shown in Table ~\ref{tab:cmse}, SolarBoost significantly reduces the RMSE compared to AverageGrid, demonstrating its ability to capture the nuances of grid-level capacity variations.



\begin{table}[t]
\vspace{-.4cm}
\caption{RMSE for generalized capacity (normalized by total generalized-capacity) of different methods on toy datasets.}
\vspace{-.4cm}
\begin{center}
\begin{tabular}{l c c}
\hline
\textbf{}&\textit{Kalman}&\textit{AR}\\
\hline

\textbf{SolarBoost}& \textbf{0.3489}& \textbf{1.097}\\
AverageGrid& 1.7879& 3.0683\\

\hline
\end{tabular}
\label{tab:cmse}
\end{center}
\end{table}

\subsubsection{Unit Capacity Output Prediction Accuracy}
\label{subsubsec:Q2}
To evaluate SolarBoost's ability to model unit output, we assess how well \(f(\boldsymbol{x_{t,i}}, \boldsymbol{\theta})\) approximates the function in Equation \eqref{eq:function}. Using inputs \(\{\boldsymbol{x}_{t,1,d}\}\) and outputs \(\{y_{t,1}\}\) from the first grid, we compare SolarBoost against IdealFit and AverageGrid. FlattenGrid is excluded, as it does not provide an estimation for this metric. Given its consistent input and output dimensions, AverageGrid serves as a relevant baseline.
Table \ref{tab:unitrmse} shows that AverageGrid's RMSE is 18.5 times higher than IdealFit's, highlighting its limitation. In contrast, SolarBoost's RMSE is only slightly higher than IdealFit's, demonstrating its superior fitting performance. 



\begin{table}[t]
\vspace{-.4cm}
\caption{Statistics for unit output of different methods on first gird of \textit{AR} dataset.}
\vspace{-.4cm}
\begin{center}
\begin{tabular}{l c c c c}
\hline
&RMSE &max &min &mean\\
\hline

IdealFit & \textit{0.0131} & 2.7086 & 0.0660 & 1.2918\\
\textbf{SolarBoost} & \textbf{0.0465} & 2.8412 & 0.0121 & 1.2913\\
AverageGrid & 0.2425 & 1.8468 & 0.6993 & 1.2922\\

\hline
\end{tabular}
\vspace{-.4cm}
\label{tab:unitrmse}
\end{center}
\end{table}


\subsubsection{Aggregated Output Prediction Accuracy}
\label{subsubsec:Q3}
We report RMSE for aggregated output prediction on the testing data in Table \ref{tab:toyrmse}. From the table we see that SolarBoost provides the lowest RMSE, which is about 20\% of that of the other methods on \textit{Kalman} dataset and 45\% on \textit{AR} dataset.




\begin{table}[t]
\caption{RMSE($10^{-2}$) for aggregated output of different methods on toy datasets.}
\vspace{-.4cm}
\begin{center}
\begin{tabular}{l c c}
\hline
 & \textit{Kalman} & \textit{AR} \\
\hline

\textbf{SolarBoost} & \textbf{1.3767} & \textbf{3.2318} \\
AverageGrid & $5.9964$ & $9.0062$ \\
FlattenGrid & $7.4832$ & $7.0471$ \\

\hline
\end{tabular}
\vspace{-.5cm}
\label{tab:toyrmse}
\end{center}
\end{table}



\subsection{Real-world Data}\label{sec:real_data_main}
\subsubsection{DPV Power Datasets}\label{sec:real_data}
We collect the output power data of DPV systems, along with corresponding numerical weather prediction (NWP) data, for five cities (City A to City E) in eastern China. This data spans the period from January 30, 2020, to September 30, 2024. The power data is recorded at 15-minute intervals (96 records/day) and measured in megawatts (MW), while NWP data has a 0.1-degree spatial resolution. 
We use three key NWP features in prediction, which have a direct and significant impact on the power generation of PV systems: irradiation, temperature, and cloud cover.
Apart from weather information, we also generate four time-based features, including sine and cosine transformations of the second of the day capturing daily and seasonal patterns, respectively. 
We aggregate input data into appropriately scaled resolution grids using averaging method.
Monthly total installed PV capacity per city was obtained from the power supply authority. We assume the capacity of all grids remain constant within each day according to the reality. 
Days with missing values in either the power output or NWP features were excluded from the dataset. The training set comprises data from January 30, 2020, to August 31, 2024, with data sizes of 85920, 82080, 85920, 84480, and 85920 for cities A through E, respectively. The testing set includes data from September 1, 2024, to September 30, 2024, with a uniform data size of 2880 for all cities.
Following standard practice for time series forecasting, we use a single temporal train-test split, and the evaluation across five cities strengthens the generalizability of our results.

\subsubsection{Performance Evaluation}
Table~\ref{tab:realrmse} presents the RMSE for SolarBoost and baseline methods on a real-world dataset across five cities (A, B, C, D and E). As demonstrated, SolarBoost consistently achieves the lowest RMSE across all locations, indicating its superior predictive accuracy in distributed photovoltaic (DPV) power output forecasting. For instance, in City A, SolarBoost reduces RMSE by 14\% compared to AverageGrid, 17\% to FlattenGrid, 15\% to CNN-LSTM, 34\% to Ave-SARIMAX and 45\% to Flat-SARIMAX.





\begin{table}[t]
\vspace{-.3cm}
\caption{RMSE($10^{-2}$) of methods on real-word dataset.}
\vspace{-.4cm}
\centering
\begin{tabular}{l c c c c c c}
    \hline
    & A & B & C & D & E \\
    \hline
    \textbf{SolarBoost}       & \textbf{3.8296} & \textbf{4.0627} & \textbf{3.9545} & \textbf{4.5823} & \textbf{4.3257} \\
    AverageGrid      & 4.4464 & 4.3217 & 4.2046 & 5.0278 & 4.6971 \\
    FlattenGrid      & 4.6276 & 4.1840 & 4.2405 & 5.1035 & 4.5524 \\
    CNN-LSTM       & 4.5147 & 4.2560 & 4.9743 & 5.0783 & 4.6883 \\
    Ave-SARIMAX  & 6.7215 & 5.1204 & 6.1438 & 6.7058 & 6.5956 \\
    Flat-SARIMAX   & 6.9783 & 5.3923 & 5.7765 & 6.5593 & 7.0519 \\
    \hline
\end{tabular}
\label{tab:realrmse}
\vspace{-0.6cm}
\end{table}





\subsubsection{Cases Study}
For AverageGrid, all grids share the same generalized capacity, equally weighting each grid's NWP features. However, since city layouts are often irregular with capacities concentrated centrally, this uniform weighting causes marginal grids to disproportionately affect overall predictions, leading to inaccuracies. FlattenGrid struggles to capture the relationship between NWP features and power output due to capacity changes, resulting in poor prediction performance.

\begin{figure}[t]
    \centering
    \includegraphics[width=0.95\linewidth]{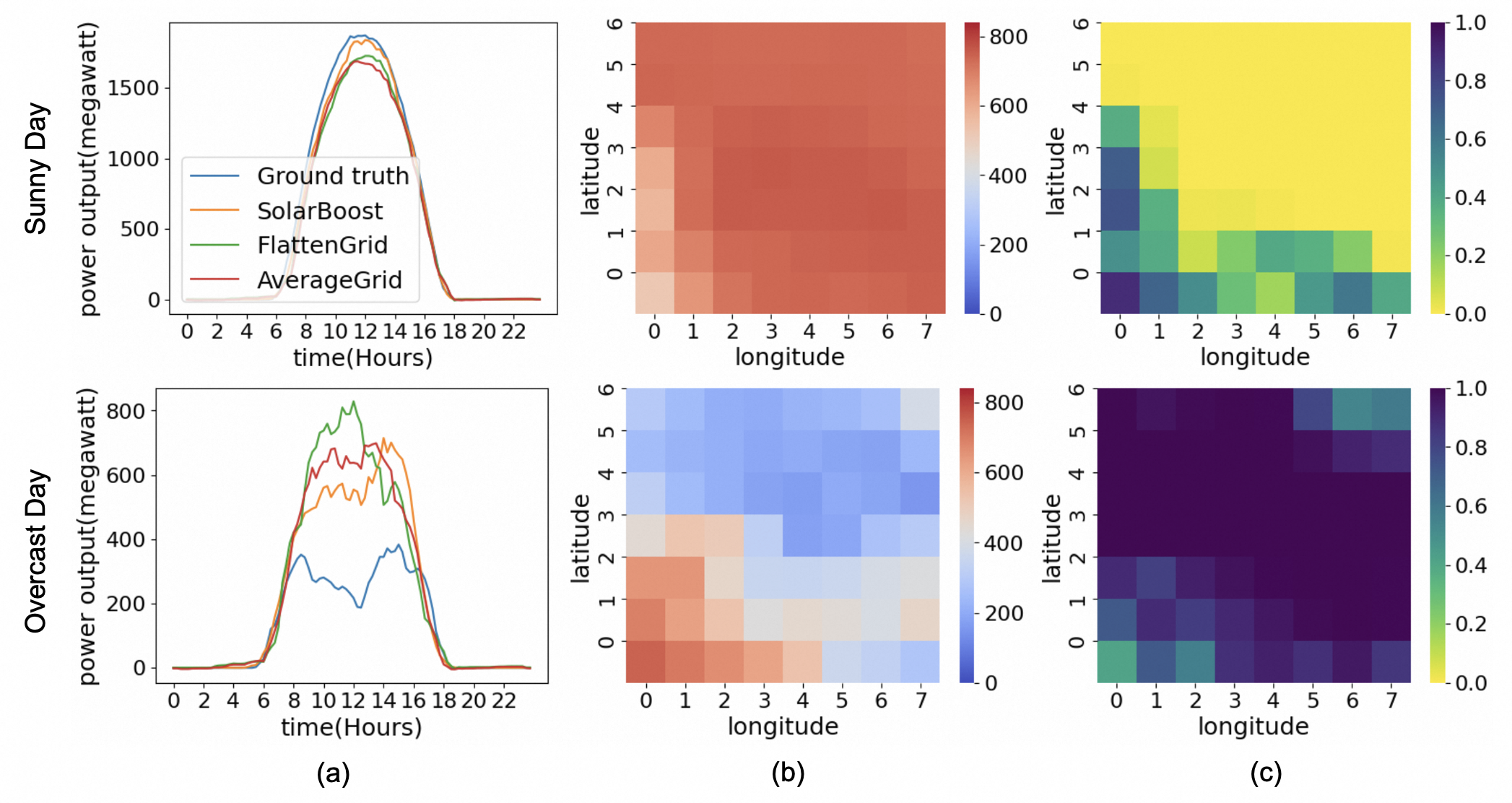}
    \vspace{-.5cm}
    \caption{
    From top to bottom, the cases are for sunny and overcast days.
    From left to right:  
    (a) Line graph comparing actual power output with model predictions.  
    (b) Heatmap of irradiation at 12 PM.  
    (c) Heatmap of cloud cover at 12 PM.}
    \label{fig:sunny case}
    \vspace{-.4cm}
\end{figure}

To demonstrate SolarBoost's effectiveness, we analyzed sunny and overcast day cases, as shown in Figure \ref{fig:sunny case}. On a sunny day with high irradiation and minimal cloud cover, baseline methods underestimated power output. AverageGrid's equal weighting caused marginal grids, especially those in the lower-left with lower irradiation and higher cloud cover, to skew NWP feature levels, while FlattenGrid failed to capture essential nuances. On an overcast day with low irradiation and high cloud cover, baseline methods overestimated power output for similar reasons. In contrast, SolarBoost provided more precise power predictions in both cases. Additionally, SolarBoost accurately tracked power output fluctuations between 12:00 PM and 2:00 PM on the overcast day, aligning well with true output, whereas AverageGrid showed minimal changes and FlattenGrid displayed incorrect inverse trends.

\subsubsection{Ablation Study}

We conduct ablation studies on the number of grids ($K$) and the regularization parameter ($\lambda$) – the only parameters unique to SolarBoost – to demonstrate their necessity and our method’s stability.

\begin{figure}[t]
    \centering
\includegraphics[width=0.95\linewidth]{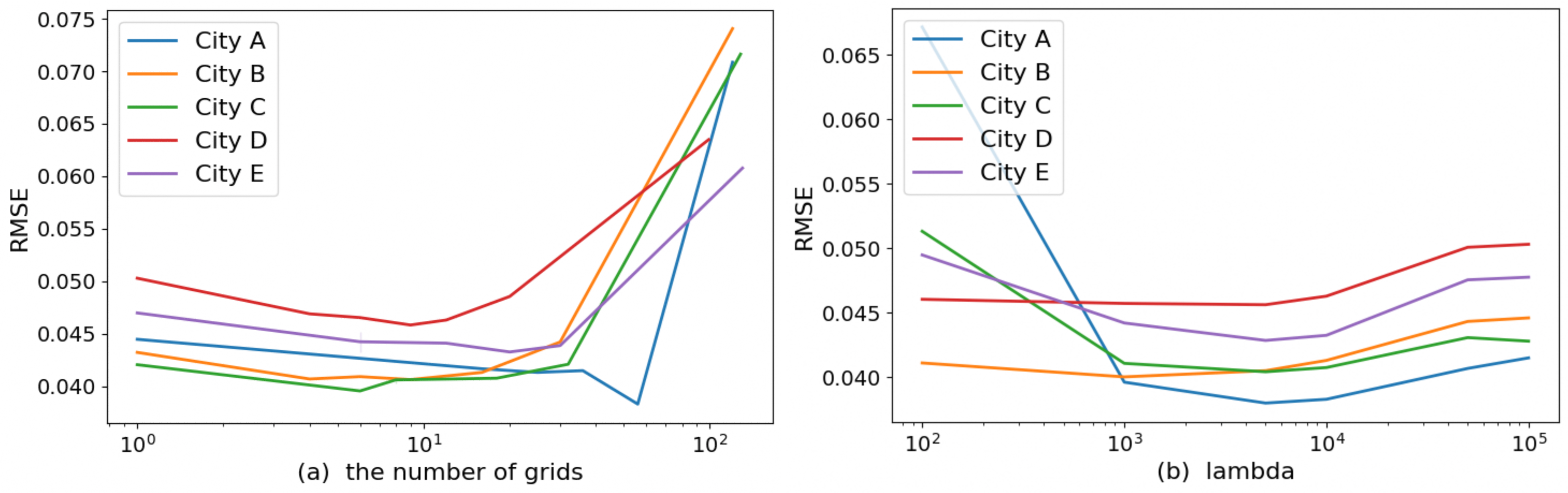}
    \vspace{-.3cm}
    \caption{RMSE vs. number of grids (a) and lambda (b) across five cities (A-E).}
    \label{fig:grids}
    \vspace{-.4cm}
\end{figure}

Figure ~\ref{fig:grids} (a) shows that RMSE consistently forms a U-shaped curve across Cities A-E (K ranging from 1-100), with performance stabilizing near the optimal K. This validates our theoretical analysis (Section~\ref{sec:thm}) regarding the advantages of grid-level modeling over aggregation, and suggests that precise K tuning is not critical.

Figure ~\ref{fig:grids} (b) shows that while RMSE is minimized at lambda = 5000 (values from 100-100,000 across Cities A-E), performance remains stable between lambda = 1000 and 10,000. The increased RMSE at lambda = 100 confirms the need for sufficient regularization to prevent overfitting.



%% file: 6.deployment.tex
\vspace{-.2cm}
\section{Deployment}\label{sec:deployment}
SolarBoost has been successfully deployed in a province in eastern China since Oct. 2024. Previously, DPV output in this region was predicted manually by experienced operators, who relied on weather information and their own empirical rules. Although this manual process could provide usable estimates, it was labor-intensive and heavily dependent on individual experience, making knowledge difficult to scale or transfer across staff. SolarBoost now  covers an area of 158,000 square kilometers, serving a population of 100 million and an installed capacity  of 37 million kilowatts (approximately 10\% of China's total distributed PV capacity in 2024).

SolarBoost substantially improves the forecast accuracy. Here, accuracy = 1 - RMSE, in line with the business stakeholders' primary performance interest. Compared to previously deployed methods, the average accuracy has improved from 95.7\% to 96.3\%, a substantial gain resulting in more efficient grid integration and management. From October 8, 2024, to January 5, 2025 (a 90-day period), SolarBoost's improved precision output reduced power curtailment by 63.89 million kilowatt-hours, generating \$0.41 million in revenue. Additionally, SolarBoost provides a valuable capacity evaluation to inform adjustments in power distribution network planning and the distributed PV system capacity planning.

\begin{figure}[t]
    \centering
    \vspace{-.1cm}
    \includegraphics[width=1\linewidth]{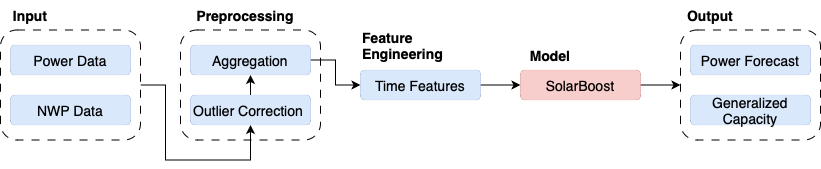}
    \vspace{-.6cm}
    \caption{SolarBoost deployment process via the eForecast platform.}
    \label{fig:deployment}
    \vspace{-.4cm}
\end{figure}

SolarBoost is currently deployed via eForecaster~\cite{Zhu2023eForecaster}, a unified forecasting platform supporting a wide range of electricity forecasting tasks, such as system load, bus load, and renewable energy (including both CPV and DPV). The platform automatically collects up-to-date historical power output and NWP data. A preprocessing module removes outliers, denoises, generates temporal features, and aggregates inputs into grids of appropriate resolution. The refined input data is then fed into SolarBoost for online training and inference.

For model training, all available historical data since January 30, 2020 are utilized, with trained models stored in a distributed file system (e.g., Aliyun OSS). The model is retrained weekly to maintain accuracy. 
For inference, day-ahead forecasts are generated every morning at 9 AM local time, and visualized via a user interface for operator usage in dispatch and reporting.

Given access to  NWP and DPV output data,  SolarBoost can be scalably deployed in other provinces.

%% file: 7.conclusion.tex
\section{Conclusions}\label{sec:conclusion}
In this paper, we introduce SolarBoost, a novel method for DPV power forecasting. We investigate the effects of capacity fluctuations and the limitations associated with neglecting them. Additionally, we formally define the DPV forecasting problem within the context of time-varying grid capacities. We then propose the SolarBoost method, which operates iteratively by alternating updates between the unit-output predictor and the generalized capacity with the objective of minimizing a surrogate loss function. Although the generalized capacity variable does not require a ground truth label, it functions as an essential latent variable in our modeling, allowing the unit-output function to better capture the complex output mechanisms and meanwhile adapting to the change of total capacity. To the best of our knowledge, SolarBoost is the first approach to explicitly incorporate generalized capacity into the model. Experiments conducted on both toy and real-world datasets demonstrate that our method effectively forecasts generalized capacities and achieves low errors in both unit capacity outputs and aggregated power predictions. The explicit operational gains achieved through deployment are enormous, particularly on the reduction in power curtailment and increased revenue. These results demonstrate both the effectiveness of technical contributions and the real-world value in actual grid management and operations.

One limitation is that SolarBoost currently relies solely on DPV data. Incorporating CPV information and transferring their characteristics to the DPV context could achieve better accuracy. Future work will also focus on developing more effective optimization algorithms, modeling capacity with greater complexity, and extending SolarBoost’s application to other renewable energy systems. 

%% file: 10.appendix.tex
\section{Detailed Derivation of Equation ~\eqref{upperbound_t_n_matrix}}
\label{appendix1}
In this appendix, we provide the detailed derivation of the matrix form of the upper bound $\mathcal{L}_t^{(n+1)}$ as expressed in equation \eqref{upperbound_t_n_matrix}.

\begin{align}
\begin{split}
&\mathcal{L}_t^{(n+1)} (\boldsymbol{\theta}, \{\boldsymbol{c}_t\}^T_{t=1})\\
&= \sum_{i=1}^{K} \frac{1}{K} \left[ -K c_{t,i} \Delta_t(\boldsymbol{x}_{t,i}; \boldsymbol{\theta}) + \boldsymbol{c}_t^{T} \boldsymbol{r}_t^{(n)}\right]^2 \\
&= \sum_{i=1}^{K} \frac{1}{K} \left[ Y_t - K c_{t,i} \Delta_t(\boldsymbol{x}_{t,i}; \boldsymbol{\theta}) - \boldsymbol{c}_t^{T} \boldsymbol{q}_t\right]^2 \\
&= \frac{1}{K} \left( K Y_t^2 + K^2 \sum_{i=1}^K c_{t,i}^2 \Delta_t (\boldsymbol{x}_{t,i}; \boldsymbol{\theta})^2 + K \boldsymbol{c}_t^T \boldsymbol{q}_t \boldsymbol{q}_t^T \boldsymbol{c}_t - \right.\\
&\quad \left. 2 K Y_t \sum_{i=1}^K c_{t,i} \Delta_t (\boldsymbol{x}_{t,i}; \boldsymbol{\theta}) - 2 K Y_t \boldsymbol{c}_t^T \boldsymbol{q}_t + 2 K \boldsymbol{c}_t^T \boldsymbol{q}_t \sum_{i=1}^K c_{t,i} \Delta_t (\boldsymbol{x}_{t,i}; \boldsymbol{\theta}) \right) \\
&= \left( Y_t^2 + K \boldsymbol{c}_t^T \boldsymbol{Q}_t \boldsymbol{c}_t + \boldsymbol{c}_t^T \boldsymbol{q}_t \boldsymbol{q}_t^T \boldsymbol{c}_t - 2 Y_t \boldsymbol{c}_t^T \boldsymbol{\Delta}_t - 2 Y_t \boldsymbol{c}_t^T \boldsymbol{q}_t + 2 \boldsymbol{c}_t^T \boldsymbol{q}_t \boldsymbol{\Delta}_t^T \boldsymbol{c}_t \right) \\
&= \left( Y_t^2 - 2 \boldsymbol{c}_t^T Y_t (\boldsymbol{\Delta}_t + \boldsymbol{q}_t) + \boldsymbol{c}_t^T (K \boldsymbol{Q}_t + \boldsymbol{q}_t \boldsymbol{q}_t^T + 2 \boldsymbol{q}_t \boldsymbol{\Delta}_t^T ) \boldsymbol{c}_t \right) \\
&= \left( Y_t^2 - 2 \boldsymbol{c}_t^T Y_t (\boldsymbol{\Delta}_t + \boldsymbol{q}_t) + \boldsymbol{c}_t^T \boldsymbol{\Sigma}_t \boldsymbol{c}_t \right) \\
&= \left\| \boldsymbol{\Sigma}_t^{\frac{1}{2}} \boldsymbol{c}_t - Y_t \boldsymbol{\Sigma}_t^{-\frac{1}{2}}(\boldsymbol{\Delta}_t + \boldsymbol{q}_t) \right\|_2^2
\label{upperbound_t_n_matrix}
\end{split}
\end{align}
where $$\boldsymbol{q}_t = \left[ f(\boldsymbol{x}_{t,1}; \boldsymbol{\theta}^{(n)}), \ldots, f(\boldsymbol{x}_{t,K}; \boldsymbol{\theta}^{(n)}) \right]^T$$ 
and
$$\boldsymbol{\Delta}_t =[\Delta_t(\boldsymbol{x}_{t,1}; \boldsymbol{\theta}), \ldots, \Delta_t(\boldsymbol{x}_{t,K}; \boldsymbol{\theta})]^T$$
represent the vectors of the aggregated predictions from the first $n$ base learners and the predictions of the $(n+1)th$ base learner for all grids at a given time step $t$, respectively.
And $$\boldsymbol{\Sigma}_t = K \boldsymbol{Q}_t + \boldsymbol{q}_t \boldsymbol{q}_t^T + \boldsymbol{q}_t \boldsymbol{\Delta}_t^T + \boldsymbol{\Delta}_t \boldsymbol{q}_t^T.$$
is a symmetric positive definite matrix. Here,
$$\boldsymbol{Q_t} = \mathrm{diag}([\Delta_t (\boldsymbol{x}_{t,1}; \boldsymbol{\theta})^2, \ldots, \Delta_t (\boldsymbol{x}_{t,K}; \boldsymbol{\theta})^2])$$ 
is a diagonal matrix constructed using the squared predictions of the $(n+1)th$ base learner.

\section{Proof for Section \ref{sec:thm}}
\subsection{Proof of Theorem \ref{thm:sup}}
\begin{proof}
    For aggregate modeling, Observation~\ref{obs:flattening} implies that there is a distribution shift on $c$ between time step $t$ and $t-1$. Let $r:C_{t}/C_{t-1}$, the learning error at time $t$ would be:
\begin{align*}
    &\mathbb{E}|rf_{agg}(\bm x') - f_\text{true}(\bm x)|\\
    =& \mathbb{E}|r\sum_{i=1}^K c_{t-1,i}\cdot \hat f(\bm x_i) - \sum_{i=1}^K c_{t,i}\cdot f(\bm x_i)|\\
    \leq& \mathbb{E}[|r\sum_{i=1}^K c_{t-1,i}\cdot \hat f(\bm x_i) - r\sum_{i=1}^K c_{t-1,i}\cdot f(\bm x_i)|\\&+ |r\sum_{i=1}^K c_{t-1,i}\cdot f(\bm x_i)- \sum_{i=1}^K c_{t,i}\cdot f(\bm x_i)|]\\
    =& C_{t}\mathbb{E}|\hat f(\bm x_i)-f(\bm x_i)|+\mathbb{E}|f(\bm x_i)|(\sum_{i=1}^K|c_{t-1,i}r-c_{t,i}|).
\end{align*}
Since the variance of $f$, $\sigma_f^2:=\mathbb{E}| f(\bm x_i)-\bar f(\bm x_i)|^2=\mathbb{E}|\hat f(\bm x_i)-f(\bm x_i)|^2+\mathbb{E}|\hat f(\bm x_i)-\bar f(\bm x_i)|^2$, $\mathbb{E}|\hat f(\bm x_i)-f(\bm x_i)|^2$ is upper bounded by $\sigma_f^2$.

Now, we bound $\sum_{i=1}^K|c_{t-1,i}r-c_{t,i}|$
\begin{align*}
    &\sum_{i=1}^K|c_{t-1,i}r-c_{t,i}|\\
    = & \sum_{i=1}^K |c_{t-1,i}C_t-c_{t,i}C_{t-1}|/C_{t-1}\\
     = &\sum_{i=1}^K |(c_{t-1,i}-c_{t,i})C_t-c_{t,i}(C_{t-1}-C_t)|/C_{t-1}\\
     \leq & K(\epsilon C_t+c_{t,i}K\epsilon)/C_{t-1}\\
     \leq & K(\epsilon C_t+C_t K\epsilon)/C_{t-1}\\
     \leq & r\epsilon K (1+K).
\end{align*}

Taking square over the original learning error, the inner product of two sources of error vanishes due to independence. Thus we have
\begin{align}
\label{eq:b1}
    \mathbb{E}|rf_{agg}(\bm x') - f(\bm x)|^2\leq C_t^2\sigma^2_f+r^2\epsilon^2 K^2(1+K)^2 M^2.
\end{align}

Similarly, for grid-level modeling, the learning error at time $t$ would be:
\begin{align*}
    &\mathbb{E}|f_\text{ours}(\bm x_{i}) - f_\text{true}(\bm x)|\\
    =& \mathbb{E}|\sum_{i=1}^K \hat c_{t,i}\cdot \hat f(\bm x_i) - \sum_{i=1}^K c_{t,i}\cdot f(\bm x_i)|\\
    \leq& \mathbb{E}[|\sum_{i=1}^K \hat c_{t,i}\cdot( \hat f(\bm x_i) - f(\bm x_i))|\\&+ \mathbb{E}|(\sum_{i=1}^K \hat c_{t,i}-\sum_{i=1}^K c_{t,i})\cdot f(\bm x_i)|],
\end{align*}
where the first term is still upper bounded by $C_t^2\sigma_f^2$, and the second term has a better control on the bias of $c_{i,t}$, which is:
\begin{align}
    \mathbb{E}(\sum_{i=1}^K \hat c_{t,i}-\sum_{i=1}^K c_{t,i})^2\leq\frac{K\sigma_c^2}{N}
    \label{eq:b2}
\end{align}
where $N$ is the sample size for each $c_{t,i}$ and $\sigma_c^2$ is the variance of $\{c_{t,i}\}_{i=1}^K$.

In the worst case, for the first term of the squared learning error, $C_t^2 \mathbb{E}|\hat f(\bm x_i)-f(\bm x_i)|^2$, its supremum  gap  between aggregate modeling and grid-level modeling is $C_t^2\sigma_f^2$. Combine bound \eqref{eq:b1} and \eqref{eq:b2}, we have 
\begin{align}
    &C_t^2\sigma_f^2 + \frac{K\sigma_c^2M^2}{N}<r^2\epsilon^2 K^2(1+K)^2 M^2 \\
    \Rightarrow &N>\frac{K\sigma_c^2M^2}{r^2K^2(1+K)^2\epsilon^2 M^2-C_t^2\sigma_f^2}
\end{align}
\end{proof}

\subsection{Proof of Theorem \ref{thm:var}}
\begin{proof}

\begin{align*}
    \mathbb{E}|\hat f'(\bm x_t)-f'(\bm x_t)|^2 = Var( f(\bm x_t)) + (\mathbb{E}\hat f'(\bm x_t)-\mathbb{E}f'(\bm x_t))^2 
\end{align*}
For a well-specified $f'$, $Var( f(\bm x_t))$ dominates the learning error. 
\begin{align*}
    Var(f'(\bm x_t)) &= Var(\sum_i c_{t,i} f(x_{t,i})) \\
    &= \sum_iVar( c_{t,i} f(x_{t,i}))+2\sum_{i\neq j} Cov (c_{t,i} f(x_{t,i}),c_{t,j} f(x_{t,j}))\\
    &\propto \sum_iVar( c_{t,i})Var( f(x_{t,i}))\\
    &\approx Var(c_{t,i}) (\frac{\partial f(x)}{\partial x}|_{x=x_0})^2 Var(x_{t,i})\\
    &\propto Var(c_{t,i})\sigma_x^2.
\end{align*}
for a point $x_0$ at which $f$ is derivable.

Given fixed $Var(f'(\bm x_t))$, we have
\begin{align}
    Var(c_{t,i}) \propto \frac{1}{\sigma_x^2}.
\end{align}
    
\end{proof}

\section{Synthetic Data Generation Process}
\label{app:toy_data}
Each element in the input tensor $\{x_{t,i,d}\}$ is independently generated from a uniform distribution $U(0,1)$.
Regarding the capacity matrix, we generate a time series of length 300 with 15 dimensions, repeating each row of this series 96 times. This repetition assumes that within any 96-length subsequence, the capacity remains constant, simulating real-world scenarios where daily capacity does not change.
We begin by generating 15 random initial values from a uniform distribution \( U(0,1) \) for each grid. To evaluate the models' overall ability to capture generalized capacity effects, we employ two distinct methods: the AR(1) model and the Kalman filter.
In the AR(1) model, each dimension is represented as an independent AR(1) process, with the coefficient and perturbation terms independently drawn from normal distributions \( N(1, \sigma^2) \) and \( N(0, \sigma^2) \), respectively.
For the Kalman filter method, dimensions are correlated through a process noise covariance matrix, defined with elements \( q_{i,j} = \sigma^2 \times 0.9^{|i-j|} \), reflecting a decaying correlation structure. Both the observation matrix and the state transition matrix are identity matrices, while the measurement error covariance is set to \( 0.1 \times \sigma^2 \).
In both methods, \( \sigma \) is set to 0.01.

\section{Implementation Details}
\label{app:implementation}
All experiments were conducted using Intel(R) Xeon(R) Platinum 8163 CPU @ 2.50GHz with 176 GB of RAM and a NVIDIA Tesla V100-SXM2-16GB GPU with 16 GB of VRAM. The operating system was Ubuntu 22.04. All models were implemented using Python 3.9 with the NumPy 1.22.0, SciPy 1.8.0, scikit-learn 1.1.3, pandas 1.4.4, and pytorch 1.11.0.

We employed a systematic approach to hyperparameter selection to ensure a fair and comprehensive comparison across all methods. For each method, we performed a grid search on a dedicated validation set, selecting hyperparameters that minimized the RMSE on this set. The validation set comprised the last 20\% of the training data held out temporally.

For SolarBoost, the regularization parameter $\lambda$ was tuned using values ranging from [10-100,000] and was ultimately set to $10$ for toy experiments and 1,000 for real-world experiments. This difference reflects the different scales and complexities of these datasets, with the real-world data requiring stronger regularization to prevent overfitting.

Regarding the boosting process parameters (number of boosting iterations and learning rate), which are shared across all methods, we maintained consistent values for both toy and real-world experiments to ensure a fair comparison. Based on preliminary experiments and common practices in boosting [Cite a paper on boosting best practices, if possible], we explored a range of values and found that setting the number of boosting iterations to 1000 and employing a learning rate of 0.01 yielded a good balance between performance and computational cost.
Limiting the maximum tree depth to 3, which was tuned across [1-10], yielded the best overall performance for all methods, as measured by the validation RMSE. We observed that deeper trees (depths of 4 or 5) tended to overfit the training data, leading to reduced generalization performance on the validation set.

For the CNN-LSTM, we used a network architecture consisting of four ConvLSTM cells. The learning rate was set to 1e-4 using the Adam optimizer, and the batch size was 5. The network was trained for 10 epochs with early stopping based on the validation MSE loss. The SXAIRME model was implemented using pmdarima-2.0.4 in Python and configured with an order parameter of (1, 0, 1) and a seasonal\_order parameter of (0, 0, 0, 96). The model was trained using maximum likelihood estimation.

\section{Supplementary Experimental Results}

\subsection{Computational Scalability}
\label{app:Computational}
This work significantly reduces the computational complexity of SolarBoost by incorporating a Kalman filter, decreasing it from $\mathcal{O}(T^3K^3)$ to $\mathcal{O}(TK^3)$.  Without the Kalman filter, simultaneous optimization would require inverting a $(TK \times TK)$ matrix.  For example, with K=50 and T=100, a single least squares update takes 213 seconds, whereas the Kalman filter-based update completes in under 1 second.

Table ~\ref{tab:cost_comparison} presents a computational cost comparison with other methods. With $T=855$, sequence length = 96, $K=56$, $D=7$, and a batch size of 5, the convergence times are 46.31 s for SolarBoost, 1765.78 s for CNN-LSTM, 2143.60 s for SARIMEX (FlattenGrid), and 1577.00 s for SARIMEX (AverageGrid). The total deployment time is under half an hour, which is acceptable for this application.

\begin{table}[t]
\centering
\caption{Computational Cost Comparison.}
\begin{tabular}{ll}
\hline
        & Convergence Time (s) \\ \hline
SolarBoost     & 46.31                \\
CNN-LSTM       & 1765.78              \\
flat-SARIMEX & 2143.60              \\
Ave-SARIMEX & 1577.00              \\ \hline
\end{tabular}
\label{tab:cost_comparison}
\end{table}

\subsection{Extended Results for Synthetic Data Experiments}
\begin{figure}[t]
\vspace{-.3cm}
    \centering
    \includegraphics[width=1\linewidth]{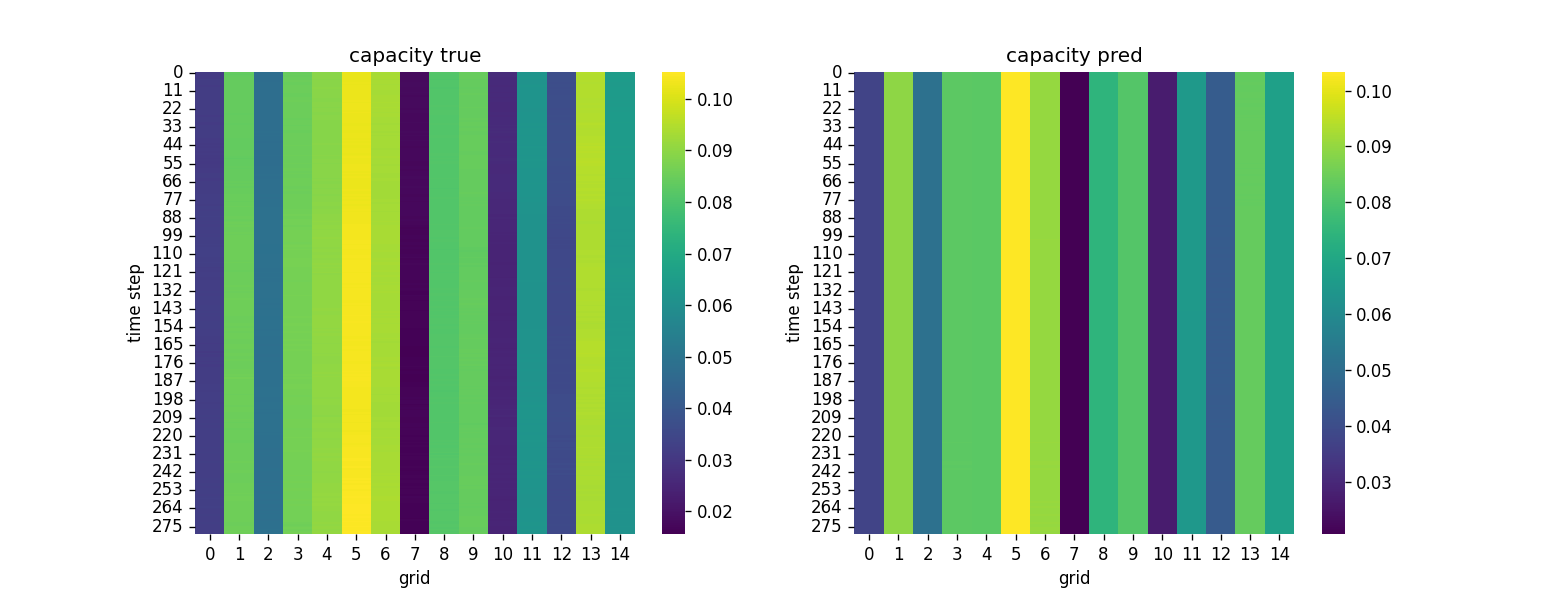}
    \vspace{-.6cm}
    \caption{Comparison of true (left) and predicted (right) generalized capacity matrices (normalized by total generalized-capacity)  for the \textit{Kalman} using heat maps. Each subfigure displays capacity values across time steps (vertical axis) and grid (horizontal axis).}
    \label{fig:capacity}
\end{figure}

To further validate SolarBoost's advantages in capturing generalized capacity (as discussed in the response to Q1 in Section~\ref{subsubsec:Q1}), Figure ~\ref{fig:capacity} illustrates a comparison between the predicted and true generalized capacities. The visual alignment of color patterns highlights that SolarBoost provides accurate estimates, demonstrating its ability to model the spatio-temporal variations in capacity.

Building upon the results for unit output prediction (related to Q2 in Section~\ref{subsubsec:Q2}), Figure \ref{fig:Kalman aggregated prediction} further illustrates that AverageGrid has a limited predictive range (an excessively high minimum and low maximum predictions), while SolarBoost accurately captures detailed variations. This reinforces the conclusion that SolarBoost excels at predicting unit output.

Supporting the aggregated output findings (Q3 in Section~\ref{subsubsec:Q3}), we find that the aggregated output values of the FlattenGrid model tend to be high at almost all points, while the AverageGrid model exhibits excessive errors at certain points, as illustrated in Figure \ref{fig:Kalman aggregated prediction}. This provides additional evidence that SolarBoost is better suited for predicting accurate aggregated output.

\begin{figure}[t]
    \centering
    \includegraphics[width=0.85\linewidth]{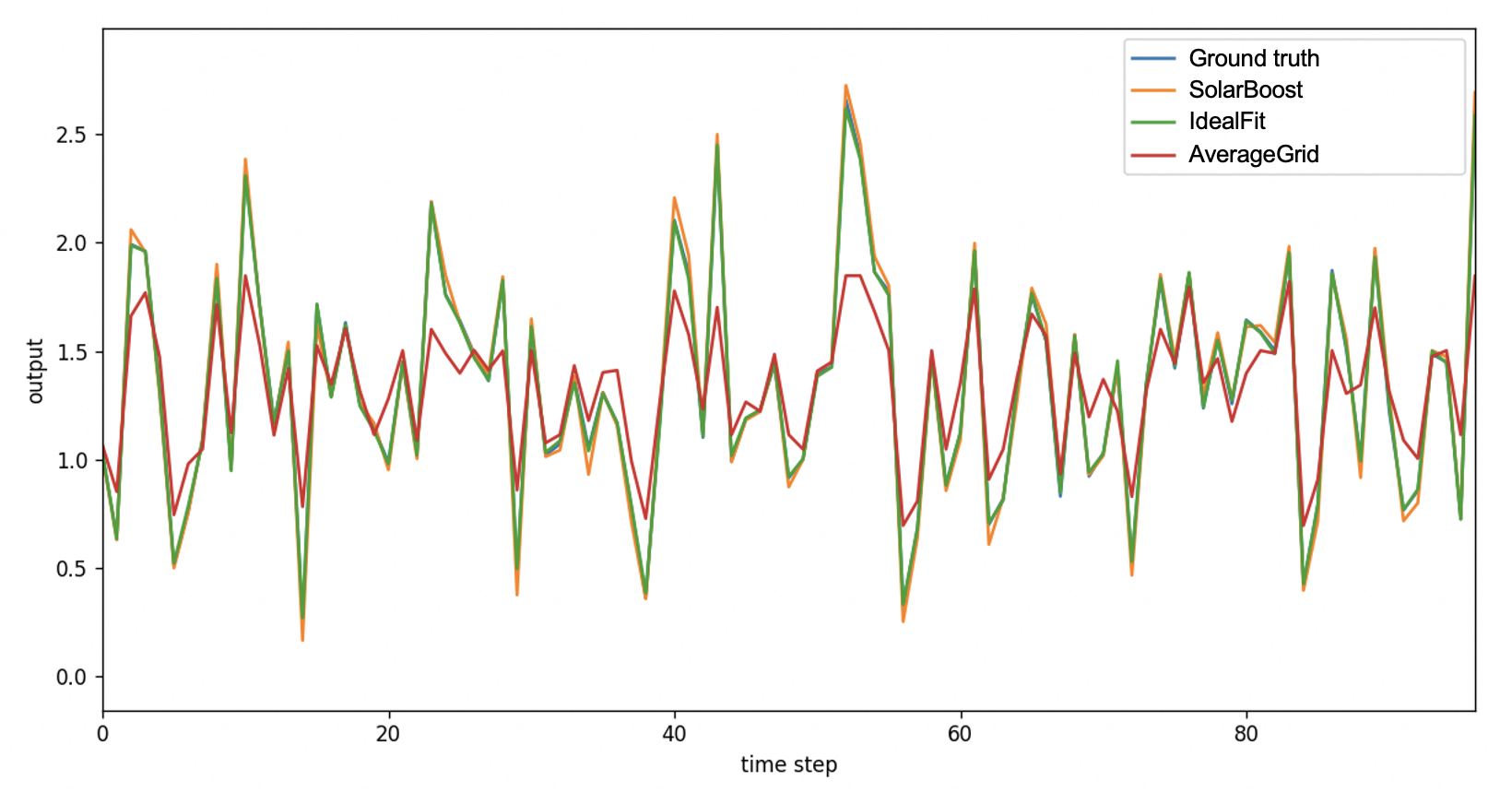}
    \vspace{-.4cm}
    \caption{
    Ground truth and different model predictions of unit capacity output for the first gird in \textit{AR} dataset.}
    \label{fig:Kalman aggregated prediction}
\end{figure}

\begin{figure}[t]
    \centering
    \includegraphics[width=0.85\linewidth]{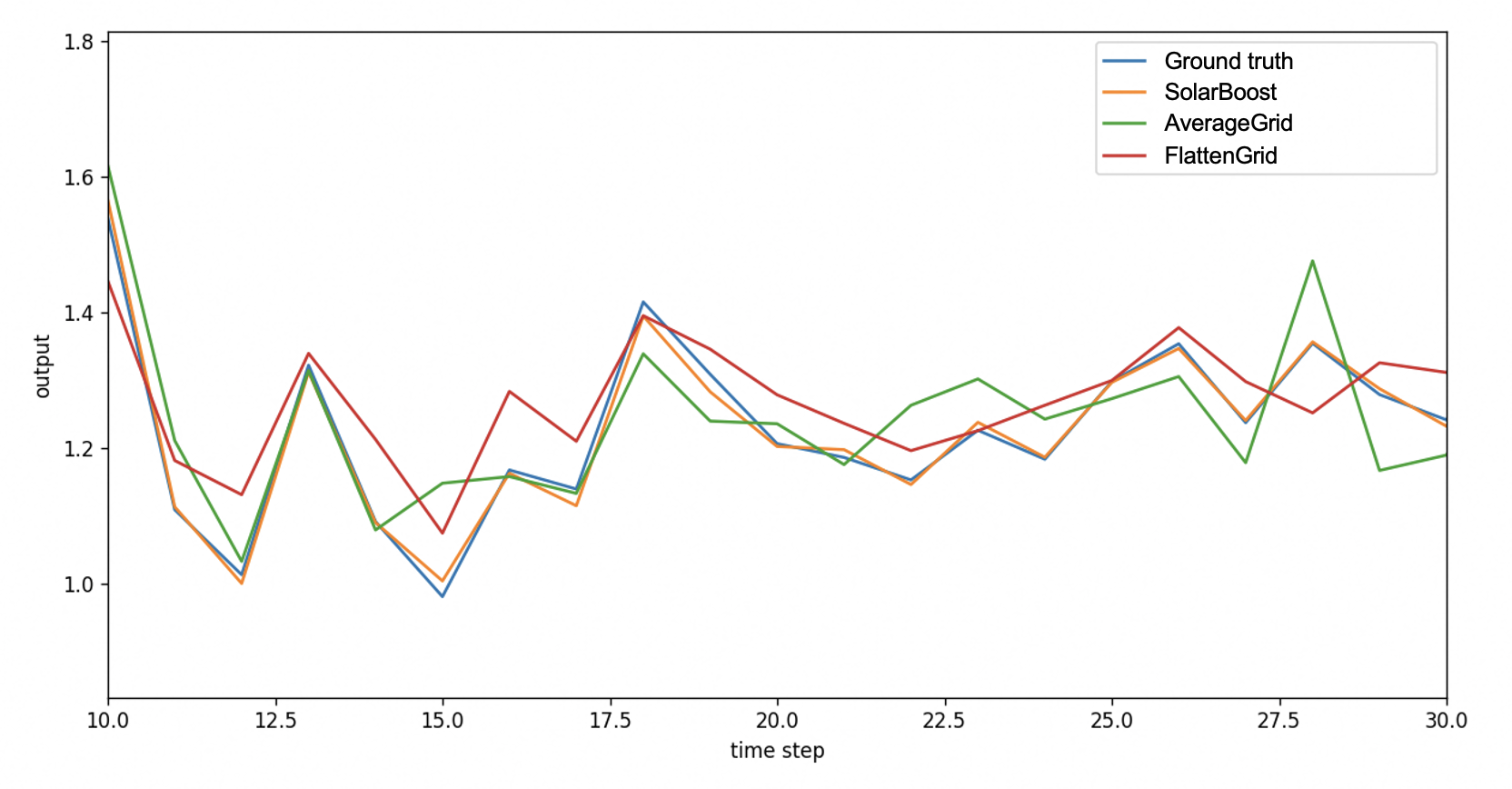}
    \caption{Ground truth and different model predictions of aggregated output for the \textit{Kalman} dataset.}
    \label{fig:Kalman aggregated prediction}
\end{figure}



%% file: 2plus.related_work.tex
\section{Related Work}
\label{app:related_work}

Most photovoltaic (PV) power forecasting research relies on weather data, such as solar irradiance, temperature, and cloud cover. Consequently, PV power forecasting primarily focuses on (1) enhancing the accuracy of weather predictions \cite{2021DeepIrradiance, 2024MLIrradiance} and (2) developing methods to transform weather predictions into PV power output, often employing machine learning (ML) and deep learning (DL) models, including 
Gradient Boosted Decision Trees (GBDT) \cite{PERSSON2017423}, 
Extreme Gradient Boosting (XGBoost) \cite{TORRESBARRAN2019151}, 
Random Forest (RF) \cite{AHMAD2018465}, 
Artificial Neural Networks (ANN) \cite{CERVONE2017274}, 
Support Vector Regression (SRV) \cite{en10070876, abuela2017solarpowerforecastingusing}, 
Convolutional Neural Networks (CNN) \cite{KORKMAZ2021117410, QU2021120996},
Long Short-Term Memory networks (LSTM) \cite{su141711083, AHMED2022115563, QU2021120996, CAO2023128669} and 
Temporal Convolutional Networks (TCN)\cite{10017379}, and so on. 
Hybrid methods \cite{KORKMAZ2021117410, QU2021120996, ZHENG2023113046, TANG2022112473, MUBARAK2023134979,10017379} and ensemble methods \cite{CAO2023128669, su141711083, AHMED2022115563, CERVONE2017274} are also commonly used to further improve forecasting accuracy.

Driven by the historical development of centralized photovoltaic (CPV) systems, initial PV power forecasting research predominantly focused on CPV scenarios. Consequently, in distributed photovoltaic (DPV) contexts, methods were often adapted from CPV forecasting techniques. For example, \cite{shi2022four, ShaoYinchi2021} transformed CPV forecasting results to obtain DPV forecasts. Alternatively, \cite{ZHAO2021120026,ZHEN2021120908} forecast each DPV plant individually and sum the results for a regional forecast. Similarly, \cite{WANG2024123890, LAI2024129716} utilizes a representative plant methodology, first forecasting the power output of select, representative DPV plants within distinct sub-regions, and then scaling or aggregating these sub-regional forecasts to obtain a final forecast for the entire distributed PV region. Despite these adaptations, the underlying approach often treats DPV systems as a collection of independent power plants, similar to CPV, failing to address the unique challenges of DPV outlined in the introduction~\ref{sec:intro}.

To the best of our knowledge, there is limited research exploring the impact of dynamic capacity on forecast accuracy.

Boosting frameworks have been widely applied in scenarios involving missing values, including fields such as biology \cite{wang2010boosting}, water treatment \cite{zhang2023miss}, traffic \cite{kaur2022missing}, visual object recognition in computer vision \cite{haffari2008boosting}, and named entity recognition in natural language processing \cite{haffari2008boosting}. 
These approaches handle missing values within the boosting framework by employing interpolation techniques or hidden variables. However, the missing values addressed by these methods are partially missing, which differs from our research context where capacity is nearly entirely missing.